\documentclass[lettersize,journal]{IEEEtran}
\usepackage{amsmath,amsfonts}
\usepackage{booktabs}
\usepackage{multirow}
\usepackage{algorithmic}
\usepackage{algorithm}
\usepackage{graphicx} 
\usepackage{array}
\usepackage{textcomp}
\usepackage{stfloats}
\usepackage{url}
\usepackage{bm}
\usepackage{verbatim}
\usepackage{graphicx}
\usepackage{cite}
\usepackage{bibentry}
\usepackage{amsthm}
\usepackage{mathtools}
\usepackage{subfigure}

\usepackage{color}
\usepackage{colortbl}
\definecolor{mygray}{gray}{.9}

\theoremstyle{definition}
\newtheorem{lemma}{Lemma}

\newtheorem{definition}{Definition}
\newtheorem{proposition}{Proposition}

\begin{document}

\title{Pair-Centric Graph Rewiring for Over-Squashing via Optimal Transport-Guided Communication Alignment}

\author{Yan Wang and Chuan-Xian Ren
\thanks{This work is supported by National Natural Science Foundation of China 62376291. (Corresponding author:
Chuan-Xian Ren.)}
\thanks{The authors are with the School of Mathematics, Sun Yat-sen University,
Guangzhou 510275, China (e-mail: rchuanx@mail.sysu.edu.cn).}}

\markboth{Journal of \LaTeX\ Class Files,~Vol.~14, No.~8, August~2021}%
{Shell \MakeLowercase{\textit{et al.}}: A Sample Article Using IEEEtran.cls for IEEE Journals}


\maketitle

\begin{abstract}
Message-passing neural networks (MPNNs) often struggle when task-relevant information is distributed across distant regions of a graph, since local propagation must compress remote signals through limited structural interfaces. Graph rewiring provides a structural response to over-squashing. Most existing methods rely on edge-level bottleneck scores or graph-level connectivity surrogates. With a limited rewiring budget, the key question is which pairwise communications most need structural support. This paper proposes PairAlign, a pair-centric graph rewiring framework that makes this question explicit through demand-support shortage. Specifically, PairAlign combines original-graph structural demand with current-graph finite-hop propagation support; their ratio highlights interactions whose communication demand is poorly supported by topology, and our theory shows that this score provides a computable proxy for the corresponding Jacobian-based shortage with a pair-level interpretation of over-squashing. Our theory reveals a two-sided effect of edge insertion: a new edge can create useful walks and simultaneously dilute existing normalized transition mass. Guided by this observation, PairAlign optimizes shortage to favor edge additions that alleviate over-squashing. Beyond selecting useful additions, PairAlign further introduces an Optimal Transport-guided rewiring mechanism to coordinate the finite edge budget for pair-level structural compatibility and shortage-target coverage. It formulates communication alignment between the candidate edge budget and the shortage targets, and the theory shows that this allocation covers shortage targets more broadly and effectively than a greedy-local assignment. Experiments on standard graph benchmarks show PairAlign's improvement across message-passing backbones, validating pair-level repair as an effective route for alleviating over-squashing. 
\end{abstract}


\section{Introduction}
\IEEEPARstart{M}{PNNS} have become a standard framework for learning on graph-structured data because they represent nodes, edges, and whole graphs through iterative local propagation over graph connectivity~\cite{gilmer2017neural,kipf2016semi}. The effectiveness of an MPNN depends on whether the input topology can support the transfer of task-relevant information. When useful evidence is distributed across distant regions of a graph, it must pass through successive local neighborhoods and be compressed into fixed-width hidden states. The graph can then become a communication bottleneck, with many remote signals forced through a small number of structural interfaces and consequently attenuated or lost. This over-squashing phenomenon has made graph rewiring a natural structural response, since the difficulty is governed not only by model depth but also by how the graph distributes communication support across node interactions~\cite{alonbottleneck,toppingunderstanding,black2023understanding}.


\begin{figure}
\centering
{\includegraphics[width=0.9\linewidth]{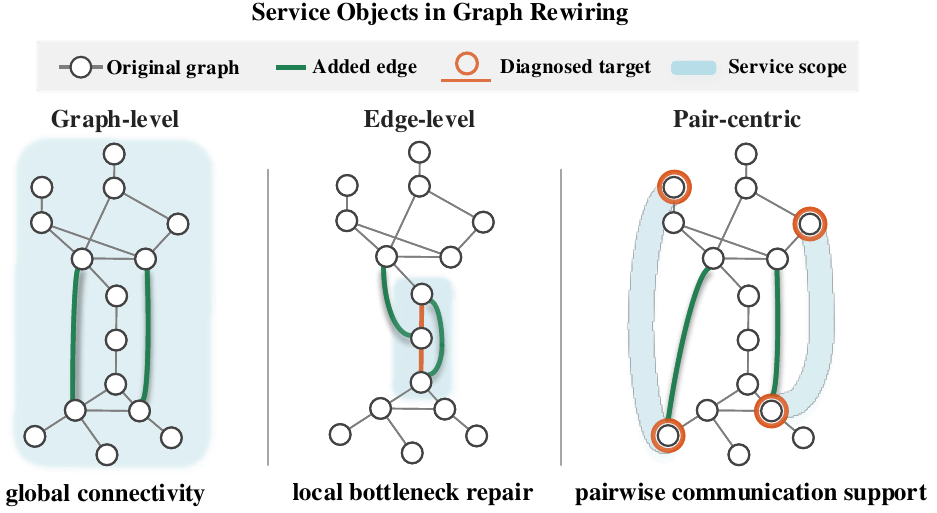}}
\caption{Different rewiring targets in graph rewiring. 
Under the same graph structure, graph-level rewiring targets global connectivity, edge-level rewiring focuses on local bottleneck repair, and PairAlign assigns rewiring support to communication-limited node pairs.}
\label{fig:service_objects}
\end{figure}

Existing rewiring methods intervene on this problem from several directions. Curvature-based and edge-centered methods use local geometric or topological cues to locate bottlenecks and modify structurally unfavorable regions~\cite{toppingunderstanding,nguyen2023revisiting}. Related local or path-aware variants enlarge the propagation structure through positional or shortest-path-based modifications~\cite{gabrielsson2023rewiring,abboud2022shortest}. A second group evaluates rewiring through graph-level communication surrogates such as spectral expansion, effective resistance, graph expansion, or differentiable global objectives~\cite{karhadkar2023fosr,black2023understanding,banerjee2022oversquashing,arnaizdiffwire}. Fig.~\ref{fig:service_objects} summarizes these two rewiring targets: graph-level methods improve aggregate connectivity, while edge-level methods repair local bottleneck structures. More recent formulations further incorporate structural priors such as locality, community structure, feature similarity, or homophily guidance into the rewiring process~\cite{barberolocality,rubiognns,linkerhagnerjoint,bi2024make,bose2025can}. These directions have broadened the design space of graph rewiring, yet they also leave a practical allocation problem under limited budgets: structural support must be assigned to the node-pair communications that need it most. This problem is not fully resolved by graph-level connectivity measures or edge-level bottleneck scores, because the scarce budget must ultimately serve interactions between node pairs. Without an explicit pair-level notion of communication shortage, rewiring can improve aggregate connectivity while leaving the most constrained pairwise communications only indirectly supported. This can weaken the effect of rewiring on the pairwise interactions most affected by over-squashing. 

To address this gap, we propose \textbf{PairAlign}, a graph rewiring framework that makes \emph{pairwise communication shortage} the basic object of structural repair. PairAlign follows the pair-level view in Fig.~\ref{fig:service_objects} and builds on the Jacobian view by prioritizing node pairs whose structural communication demand is high while their finite-hop propagation support remains limited. To make this view operational, PairAlign defines shortage as the ratio between original-graph structural demand and current-graph finite-hop propagation support. This score highlights interactions whose communication demand is poorly supported by the current topology. Our theory shows that this shortage score provides a computable proxy for the corresponding Jacobian-based shortage with a pair-level interpretation of over-squashing.

To turn pair-level diagnosis into edge additions, PairAlign scores candidate
edges by their support effect on high-shortage pairs. A useful candidate edge
need not directly connect the target pair; its value depends on the support
improvement it provides to shortage pairs. Our theory shows that the gain from
edge insertion is not always positive: a new edge can create useful walks, while
normalization reduces the transition mass assigned to existing neighbors. Thus,
PairAlign recomputes shortage during rewiring and selects edges by their current
shortage reduction. To coordinate the limited budget, PairAlign introduces an OT-guided
rewiring mechanism to allocate candidate-edge mass across shortage targets. This keeps the budget from
concentrating on a few locally cheap pairs and coordinates allocation according
to the shortage distribution over graph-wide targets. Accordingly, the main contributions of
this work are summarized as follows.
\begin{itemize}
    \item The paper introduces a \emph{pair-centric} formulation of graph rewiring based on communication shortage. The shortage score ranks node pairs by the lack of current-graph support relative to the original-graph structural demand. The theory shows the formulation makes Jacobian-based shortage tractable through graph structure and yields an interpretable pair-level diagnostic for over-squashing.

    \item PairAlign is proposed as an OT-guided graph rewiring framework for shortage-aware edge addition. Candidate edges are scored by current shortage reduction, and the OT-guided allocation coordinates the limited edge budget across shortage targets. This yields a globally coordinated rewiring strategy and avoids concentrating additions on only a few targets.

    \item We validate PairAlign on standard graph benchmarks across message-passing backbones. The results and mechanism studies show that PairAlign can effectively repair the communication shortage and improve performance. This supports pair-level repair as an effective route for alleviating over-squashing.
\end{itemize}

The remainder of the paper is organized as follows. Section~II reviews graph
rewiring methods for over-squashing. Section~III develops PairAlign from the
shortage formulation to OT-guided edge allocation. Section~IV evaluates the
method and analyzes its rewiring mechanism. Section~V concludes the paper and discusses future directions.

\section{Related Work}
Graph rewiring has become a central structural strategy for mitigating over-squashing in MPNNs. Over-squashing arises when the input topology concentrates multiple information flows through restricted structural interfaces, causing useful signals to be compressed into fixed-width node representations~\cite{alonbottleneck,toppingunderstanding}. This phenomenon links the effectiveness of message passing to graph topology, including curvature, expansion, information contraction, and effective resistance properties~\cite{banerjee2022oversquashing,black2023understanding,barberolocality}. Expander-based propagation further shows that bottleneck-resistant computational structures can improve message passing without relying only on the original input topology~\cite{deac2022expander}. Recent surveys further organize over-squashing mitigation into rewiring, spectral, curvature-based, and propagation-oriented families, confirming graph rewiring as a sustained research direction in MPNN design~\cite{akansha2025over}. Related spatial and path-aware interventions expand the computational neighborhood through positional encodings, shortest-path message passing, or discrete geometric rewiring, providing additional ways to improve propagation beyond immediate graph adjacency~\cite{gabrielsson2023rewiring,abboud2022shortest,bober2023rewiring}. In this setting, the main question is how a modified computational graph should distribute propagation support under a finite structural budget.

A major line of rewiring methods diagnoses over-squashing through local geometry and edge-level bottlenecks. SDRF uses graph curvature to locate structurally critical regions for rewiring, making curvature one of the earliest operational proxies for topology-induced over-squashing~\cite{toppingunderstanding}. BORF extends this perspective through Ollivier-Ricci curvature and connects negative curvature with over-squashing and positive curvature with over-smoothing, yielding a unified local-geometric rewriting principle~\cite{nguyen2023revisiting}. Recent studies have refined this view by examining the empirical behavior of curvature-guided rewiring and by proposing more expressive curvature variants for detecting over-squashed edges~\cite{torieffectiveness,chenrethinking}. This family establishes a strong local diagnostic foundation for graph rewiring, and it motivates a complementary question concerning which node-pair communications should receive structural support after local bottlenecks have been detected.

Another line of work formulates rewiring through graph-level communication surrogates. FoSR improves spectral expansion by adding sparse edges that increase the spectral gap, thereby strengthening global connectivity for message passing~\cite{karhadkar2023fosr}. Effective resistance-based methods use resistance as a communication proxy and guide edge modification through graph-wide improvements in connectivity~\cite{black2023understanding,shen2024graph}. DiffWire introduces a differentiable rewiring objective based on the Lovász bound~\cite{arnaizdiffwire}, and LASER studies the trade-off among connectivity, locality, and sparsity through locality-aware sequential rewiring~\cite{barberolocality}. More recently, spectrum-preserving sparsification has strengthened this family by improving connectivity while retaining sparsity and Laplacian spectral structure~\cite{liangmitigating}. Cayley graph propagation extends the expander-based perspective by propagating over complete Cayley graph structures to obtain bottleneck-free computational templates~\cite{wilson2025cayley}. These methods provide mature graph-level principles for structural improvement, and they set the stage for rewiring objectives that explicitly organize the budget around a distribution of underserved node-pair interactions.

Recent rewiring studies have also incorporated task-relevant signals beyond purely topological criteria. ComFy uses community structure and feature similarity to guide rewiring toward stronger label-community alignment~\cite{rubiognns}. JDR jointly rewires the graph and denoises node features by aligning the leading spectral spaces of graph and feature matrices~\cite{linkerhagnerjoint}. In heterophilic graphs, DHGR modifies neighborhoods by adding homophilic edges and pruning heterophilic ones, and label-guided rewiring further uses label information to improve the compatibility between graph structure and downstream classification~\cite{bi2024make,bose2025can}. These studies show that rewiring objectives benefit from structural signals that better match the role of the graph in downstream prediction.

PairAlign keeps the criterion topology-driven and uses topology to define a pair-level shortage signal that brings node-pair communication needs into the rewiring objective. It frames rewiring as pair-centric structural repair, where the target is a communication-deficient node pair whose demand exceeds the support available under the current topology. Its OT-guided allocation assigns candidate additions to high-shortage pairs under the finite budget, turning graph rewiring into a budgeted allocation problem for the pairwise communications most constrained by over-squashing.

\section{Methods}
\subsection{Preliminaries}
Let $G=(V,E,\mathbf H)$ be a graph with node set $V$, edge set $E$, and node feature matrix $\mathbf H\in\mathbb R^{n\times d}$, where $n=|V|$. Set $\mathbf H^{(0)}=\mathbf H$, and let $h_u^{(0)}$ denote its $u$-th row. The adjacency matrix of $G$ is denoted by
$\mathbf A\in\{0,1\}^{n\times n}$. For any $u,v\in V$, let $d_G(u,v)$
denote the shortest-path distance on the original graph. We consider an
add-only rewiring setting. Let $\bar E$ denote the set of candidate non-edges
of $G$. A rewiring solution is represented by a matrix
$\mathbf B\in\mathbb R_{\ge 0}^{n\times n}$. Its feasible set is
\begin{equation}
\mathcal B
:=
\bigl\{
\mathbf B:
\mathbf B=\mathbf B^{\top},\
\operatorname{diag}(\mathbf B)=\mathbf 0,\
\operatorname{supp}(\mathbf B)\subseteq \bar E
\bigr\}.
\end{equation}

For budgeted allocation, let $\mathcal T$ be the finite target-pair set, let
$\mathcal E\subseteq\bar E$ be the finite candidate-edge pool, and let
$\mathbf p\in\Delta(\mathcal T)$ be the shortage-target distribution used by the
allocation. The added-budget distribution induced by the edge-addition
variables is denoted by $\mathbf q(\mathbf B)\in\Delta(\mathcal E)$ and is
used as the source marginal over candidate added edges.

The rewired graph is encoded by
$\mathbf{W}=\mathbf{A}+\mathbf{B}.$
Let $\mathbf{D}$ be the degree matrix of $\mathbf{W}$. The row-normalized propagation matrix $\mathbf{P}$ on the rewired graph is defined as $\mathbf{P}=\mathbf{D}^{-1}\mathbf{W}.$ Let $\mathcal{P}=\{(u,v)\in V\times V:\, u\neq v\}$ denote the set of ordered node pairs. Unless explicitly specified, all pairwise quantities in the sequel are indexed over $\mathcal{P}$, and all distance-based constraints are defined on the original graph $G$.
At layer $r$, a row-normalized MPNN updates node $j$ by
\begin{equation}
\label{eq:row_normalized_mpnn}
h_j^{(r)}
=
\phi_r\left(
h_j^{(r-1)},
\sum_i P_{ij}\psi_r(h_i^{(r-1)})
\right),
\end{equation}
where $\psi_r$ is the message map and $\phi_r$ is the update map.


\subsection{Pairwise Communication Shortage as a Demand--Support View of Over-Squashing}
\textbf{Motivation.} Existing graph rewiring studies usually evaluate structural modifications through edge-level bottlenecks, local geometric indicators, or graph-level connectivity surrogates. These criteria provide useful diagnoses of unfavorable graph structures. Under a limited rewiring budget, the central decision is which node-pair interactions should receive structural support. From the perspective of over-squashing, the constrained object is the communication between distant node pairs whose information exchange must pass through limited multi-hop propagation pathways. This observation motivates a pair-centric view that allocates structural budget to node pairs whose communication demand is poorly supported by the current graph. We introduce pairwise communication shortage as a demand-support diagnostic for over-squashing risk and use it to prioritize the node-pair interactions that should be supported by rewiring.

\textbf{Formulation.} To make the pair-centric view operational, PairAlign scores over-squashing risk by comparing structural demand with propagation support for each node pair. The demand term is fixed on the original graph and reflects how much structural communication the pair requires. The support term is measured on the current computational graph and records how much finite-hop propagation is available to that pair. A large shortage marks a pair whose communication demand receives limited support from the current topology. We use this score as a rewiring diagnostic for over-squashing risk and first define the two quantities that enter it: pairwise communication demand and propagation support.

\begin{definition}[Pairwise Communication Demand]
Let $\mathcal{P}$ denote the set of ordered node pairs. 
The pairwise communication demand is the function $\omega:\mathcal{P}\to \mathbb{R}_{\ge 0}$ defined by
\begin{equation}
\label{eq:demand}
\omega(u,v)=d_G(u,v)^p,\qquad (u,v)\in\mathcal{P},
\end{equation}
where $d_G(u,v)$ is the shortest-path distance between $u$ and $v$ on $G$, and $p>0$ controls the growth rate of communication demand with respect to distance.
\end{definition}

This definition captures the structural side of pairwise communication demand. By assigning greater demand weights to distant node pairs, it brings to the foreground those interactions that rely more heavily on long-range information exchange; under multi-hop propagation, such interactions typically traverse longer transmission chains and are more vulnerable to compression caused by bottlenecks and information attenuation~\cite{alonbottleneck,toppingunderstanding}. In this sense, the shortest-path distance provides a natural topology-based prior for long-range pairwise communication demand.

We now define the support side of the demand-support diagnostic. For a finite-depth message passing model, communication from $u$ to $v$ can only use propagation walks whose length is within the effective depth of the model. PairAlign measures support by the finite-hop propagation mass induced by the current computational graph.

\begin{definition}[Pairwise Propagation Support]
\label{def:support}
Let $\mathbf{W}$ be a graph with propagation matrix $\mathbf{P}$, let $\mathcal{P}$ denote the set of ordered node pairs, and let $\alpha_1,\ldots,\alpha_K\in\mathbb{R}_{\ge 0}$ be nonnegative hop weights satisfying $\sum_{\ell=1}^{K}\alpha_\ell=1$. The finite-hop pairwise propagation support on $\mathbf{W}$ is the function $s(\,\cdot\,;\mathbf{W}):\mathcal{P}\to\mathbb{R}_{\ge 0}$ defined by
\begin{equation}
\label{eq:k-hop-prc}
s(u,v;\mathbf{W})
=
\sum_{\ell=1}^{K}\alpha_{\ell}\,[\mathbf{P}^{\ell}]_{uv},
\qquad (u,v)\in\mathcal{P}.
\end{equation}
\end{definition}

In the main method, PairAlign uses the uniform hop average $\alpha_\ell=1/K$.
Under this convention, $s(u,v;\mathbf W)$ measures the average propagation mass that can travel from $u$ to $v$ within $K$ message-passing steps. The following lemma gives the corresponding Jacobian bound.
\begin{lemma}[Support bounds normalized Jacobian influence]
\label{prop:normalized_message_passing_influence_bound}
Consider the $K$-layer MPNN defined as Eq.~\eqref{eq:row_normalized_mpnn}, and
let $h_v^{(\ell)}$ be the representation of node $v$ at layer
$\ell=0,\ldots,K$. Assume that $\psi_r$ is Lipschitz and $\phi_r$ is Lipschitz
in its propagated-message argument at each layer $r=1,\ldots,K$, and let $L_r$
denote the corresponding layer Lipschitz constant. For
every ordered pair $(u,v)\in\mathcal P$, the support
$s(u,v;\mathbf W)$ upper bounds the normalized finite-depth
Jacobian influence $\mathcal I(u,v)$ as
\begin{equation*}
\mathcal I(u,v)=\frac{1}{K}\sum_{\ell=1}^{K}
\frac{\left\|\partial h_v^{(\ell)}/\partial h_u^{(0)}\right\|}
{\prod_{r=1}^{\ell}L_r}
\le
s(u,v;\mathbf W).
\end{equation*}
\end{lemma}

At the mechanism level, over-squashing appears as weak influence between
distant nodes after finite-depth message passing. This influence reflects the
graph structure together with the MPNN model functions.
When the model functions are Lipschitz,
Lemma~\ref{prop:normalized_message_passing_influence_bound} isolates the
structural side and shows that the finite-hop propagation mass
$s(u,v;\mathbf W)$ is a structural upper bound on the normalized
Jacobian influence.

For a pair with large structural demand $\omega(u,v)$, a small value of
$s(u,v;\mathbf W)$ means that the current graph leaves little
finite-depth influence available for this pairwise message passing, which
indicates a higher risk of over-squashing. The shortage score serves as a
diagnostic to record this mismatch between structural demand and support on the
graph topology. Overall, PairAlign ranks the resulting
communication-deficient pairs by the ratio below.

\begin{definition}[Pairwise Communication Shortage]
For each ordered node pair $(u,v)\in\mathcal{P}$, let $\omega(u,v)$ denote its pairwise communication demand in Eq.~\eqref{eq:demand} and let $s(u,v;\mathbf{W})$ denote its finite-hop pairwise propagation support in Eq.~\eqref{eq:k-hop-prc}. The pairwise communication shortage on $\mathbf{W}$ is defined as the function $\mathcal{S}(\,\cdot\,;\mathbf{W}):\mathcal{P}\to\mathbb{R}_{\ge 0}$ given by
\begin{equation}
\label{eq:shortage definition}
\mathcal{S}(u,v;\mathbf{W})
=
\frac{\omega(u,v)}
{s(u,v;\mathbf{W})+\varepsilon},
\qquad (u,v)\in\mathcal{P},
\end{equation}
where $\varepsilon>0$ is a small positive constant.
\end{definition}
For fixed $(u,v)$, we abbreviate $s(\mathbf W)=s(u,v;\mathbf W)$ and $\mathcal{S}(\mathbf W)=\mathcal{S}(u,v;\mathbf W)$.
Under a fixed demand specification and propagation protocol, $\mathcal{S}(\mathbf W)$
ranks node pairs by how much structural demand remains unsupported by finite-hop
propagation; larger values indicate pairs that should receive higher priority
in shortage-aware rewiring. The Jacobian view of over-squashing relates the
problem to weak finite-depth influence between nodes. In graph rewiring, weak
influence alone is not enough to rank repair priorities, since some weakly
influenced pairs may require only limited structural communication.
A pair with large structural demand and weak normalized Jacobian influence
reflects a more severe communication shortage. We measure this shortage by the
ratio between structural demand and normalized Jacobian influence, defined below
as the Jacobian-based shortage. Proposition~\ref{prop:v5_shortage_jacobian_bottleneck}
shows that the computable shortage used by PairAlign serves as a reasonable
proxy for this Jacobian-based ratio.

\begin{proposition}[Proxy of Jacobian-based shortage]
\label{prop:v5_shortage_jacobian_bottleneck}
For a node pair $(u,v)$, suppose that the assumptions of
Lemma~\ref{prop:normalized_message_passing_influence_bound} hold. Define the
Jacobian-based communication shortage by
\begin{equation*}
\mathcal{S}^{\rm Jac}(\mathbf W)
=
\frac{\omega(u,v)}{\mathcal I(u,v)+\varepsilon},
\end{equation*}
where $\mathcal I(u,v)$ is the normalized Jacobian influence in
Lemma~\ref{prop:normalized_message_passing_influence_bound}. Then the following statements
hold.

\emph{(i) Bottleneck lower bound.}
$\mathcal{S}(\mathbf W)\le \mathcal{S}^{\rm Jac}(\mathbf W)$.

\emph{(ii) Constant-factor equivalence in the non-degenerate regime.}
Assume that the normalized Jacobian influence from $u$ to $v$ retains a $c$
fraction of $s(\mathbf W)$, i.e.,
$\mathcal I(u,v)\ge c\,s(\mathbf W)$ with
$c\in(0,1]$.
$\mathcal{S}(\mathbf W)$ and $\mathcal{S}^{\rm Jac}(\mathbf W)$ satisfy the following
constant-factor equivalence:
\begin{equation*}
\mathcal{S}(\mathbf W)
\le
\mathcal{S}^{\rm Jac}(\mathbf W)
\le
c^{-1}\mathcal{S}(\mathbf W).
\end{equation*}
\end{proposition}

The non-degenerate regime gives a sufficient setting for analyzing graph
rewiring as a structural way to alleviate over-squashing. It assumes that the
finite-hop support supplied by the graph is not erased by the model functions,
so topology changes can still improve the communication bottlenecks. With this
condition, $\mathcal{S}(\mathbf W)$ and $\mathcal{S}^{\rm Jac}(\mathbf W)$ are equivalent up to
a constant factor. A large $\mathcal{S}(\mathbf W)$ points to a node pair whose
communication demand is high while the current graph provides limited
finite-hop support. Thus, reducing the computable score $\mathcal{S}(\mathbf W)$ also
reduces the corresponding Jacobian-based shortage. This makes the rewiring
objective focus on the pairs most vulnerable to over-squashing.

Once the shortage score selects demand--support deficient pairs, we now consider how an edge insertion changes the support of those targets. PairAlign acts on the graph through candidate edges and each edge is evaluated by its support change for a shortage target under the row-normalized propagation matrix. This effect is not monotone. Adding a new neighbor creates walks through the new edge, while row normalization reallocates probability mass away from the original neighbors of the same row. To make this local effect explicit, we consider the row update $e=a\to b$ induced by the undirected candidate edge $\{a,b\}$. For $\beta\ge0$, let $\mathbf P_\beta$ denote the row-normalized propagation matrix obtained after adding edge weight $\beta$ to this row update, with $\mathbf P_0=\mathbf P$. The first-order support change $g_e(u,v)$ at the uninserted graph is
\begin{equation*}
g_e(u,v)
=
\left.
\frac{d}{d\beta}
\frac{1}{K}\sum_{\ell=1}^{K}
\left[\mathbf P_\beta^\ell\right]_{uv}
\right|_{\beta=0^+}.
\end{equation*}

\begin{proposition}[First-order support change from edge insertion]
\label{prop:bridge_service_displacement}
Assume that $b\notin N(a)$ before insertion and let $d_a$ be the original row degree of $a$. For $e=a\to b$, the first-order finite-hop support change of $(u,v)$ is
\begin{equation*}
g_e(u,v)=B_e(u,v)-D_e(u,v),
\end{equation*}
where the added-path contribution is
\begin{equation*}
\begin{aligned}
B_e(u,v)=
\frac{1}{K}\sum_{\ell=1}^{K}
\sum_{i=1}^{\ell}
[\mathbf P^i]_{ua}
\frac{[\mathbf P^{\ell-i}]_{bv}}{d_a},
\end{aligned}
\end{equation*}
and the normalization loss over the original neighbors of $a$ is
\begin{equation*}
\begin{aligned}
D_e(u,v)=
\frac{1}{K}\sum_{\ell=1}^{K}
\sum_{i=1}^{\ell}
[\mathbf P^i]_{ua}
\frac{\sum_{z\in N(a)}\mathbf P_{az}
[\mathbf P^{\ell-i}]_{zv}}{d_a}.
\end{aligned}
\end{equation*}
Here $B_e(u,v)$ sums walks that reach $a$, use the new step $a\to b$, and continue to $v$, while $D_e(u,v)$ sums the mass removed from walks that would have continued through the original neighbors of $a$. For the undirected insertion $\{a,b\}$, the first-order support change is the symmetrized quantity $g_{\{a,b\}}(u,v)=g_{a\to b}(u,v)+g_{b\to a}(u,v)$.
\end{proposition}

Proposition~\ref{prop:bridge_service_displacement} shows that an inserted edge
does not simply add propagation support to a target pair. The support increases
only when the added-path contribution exceeds the loss, namely
$B_e(u,v)>D_e(u,v)$. An edge is useful only when it lowers
shortage on the rewiring graph, not merely when it increases connectivity.
This view is consistent with evidence that more connectivity is not always
preferable~\cite{jamadandi2024spectral}. PairAlign keeps optimizing shortage
during rewiring to ensure the effectiveness of edge insertions.

\subsection{PairAlign with OT-Guided Communication Alignment for Shortage Reduction}

Under a finite edge addition budget, graph rewiring needs a global allocation rule for assigning candidate edges to high-shortage pairs. This allocation must address two coupled issues. High-shortage pairs are not independent targets. Several targets may rely on overlapping communication regions, and one candidate edge may affect multiple targets at once. If edges are selected independently, as in Greedy-Local, the budget can concentrate on targets that are easier to improve locally, leaving other shortage regions undercovered. The allocation also needs to account for how well a candidate edge can serve the target pairs it is meant to support. A candidate edge should be favored only when it can provide useful propagation support for the corresponding shortage pairs.

\begin{figure*}[!t]
\centering
\includegraphics[width=0.95\textwidth]{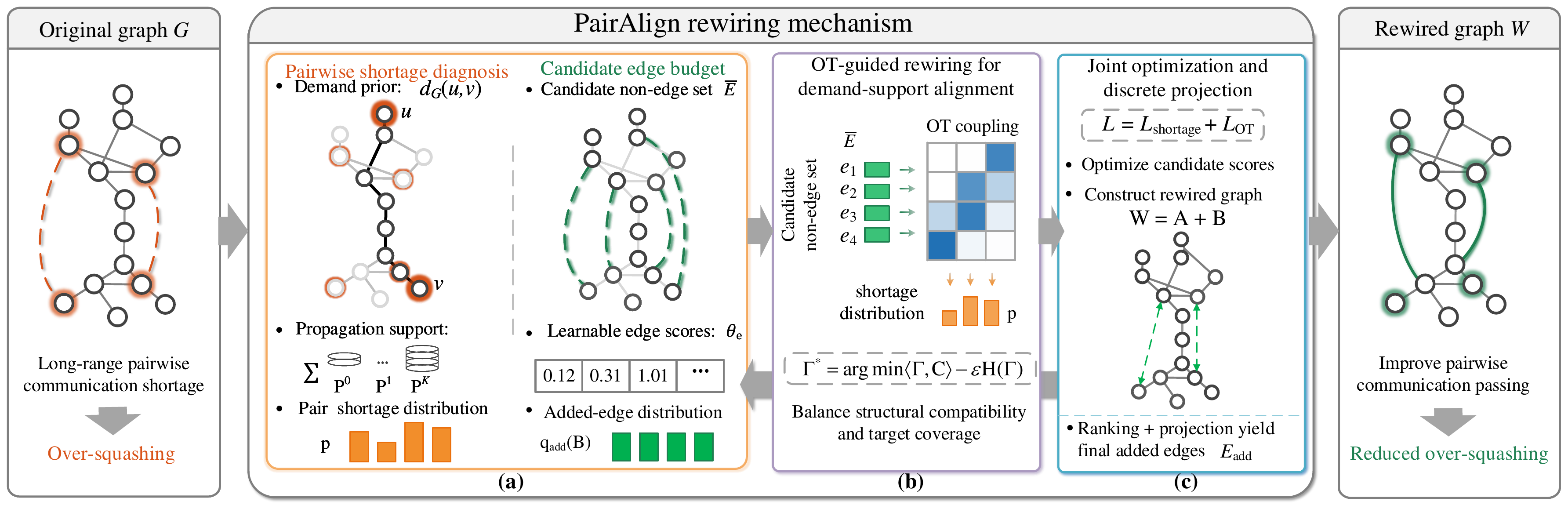}
\caption{PairAlign rewiring mechanism.
PairAlign rewires a graph by aligning limited edge additions with pairwise communication shortages. (a) The method constructs a shortage-target distribution over communication-deficient node pairs, and defines a candidate-edge budget to represent the limited structural resources available for repair. (b) OT-guided alignment solves for a coupling between candidate edges and shortage targets under a transport cost that balances structural compatibility and target coverage. (c) The optimization loop updates candidate-edge scores, re-evaluates shortage on the current rewired graph, and feeds this signal back into edge selection.
After discrete projection, the selected set $E_{\mathrm{add}}$ forms $\mathbf W=\mathbf A+\mathbf B$.
By assigning limited rewiring capacity to communication-deficient node pairs, PairAlign strengthens message passing and alleviates over-squashing.}
\label{fig:pairalign_mechanism}
\end{figure*}

To establish a pair-level rewiring mechanism that captures both budget coverage
and structural compatibility, PairAlign formulates the communication
alignment between the candidate edge budget and the shortage targets as an
Optimal Transport problem, as shown in Fig.~\ref{fig:pairalign_mechanism}. Drawing on the coupling view of Optimal
Transport~\cite{villani2009optimal,cuturi2013sinkhorn}, OT-guided alignment treats the candidate edge budget as structural resources to be assigned and pairs with high shortage as targets that require support. The transport cost measures how suitable a candidate edge
is for supporting a given shortage target. The optimal coupling gives the
allocation solution by matching the finite candidate-edge budget to the shortage
targets with minimum total transport cost. As a result, structural resources are
preferentially assigned to node pairs that both need support and can be
effectively served by the selected candidates.

\textbf{Formulation of OT-guided Alignment.}
Under this view, PairAlign aligns the added-budget distribution $\mathbf{q}(\mathbf{B})$ with the shortage distribution $\mathbf{p}$ on the original graph under a transport cost $\mathbf{C}$ that encodes their structural compatibility. The corresponding OT problem is defined as Eq.~\eqref{eq:OTalignment} and the optimal coupling $\mathbf{\Gamma}^{\star}$ characterizes the cost-compatible allocation,
\begin{equation}
\label{eq:OTalignment}
\mathbf{\Gamma}^{\star}
=
\arg\min_{\mathbf{\Gamma} \in \Pi(\mathbf{q}(\mathbf{B}),\mathbf{p})}
\;
\langle \mathbf{\Gamma},\mathbf{C}\rangle
-
\varepsilon H(\mathbf{\Gamma}),
\end{equation}
where $
\Pi(\mathbf{q}(\mathbf{B}),\mathbf{p})
=
\{
\mathbf{\Gamma}_{ij}\ge 0, \forall i,j
\mid
\mathbf{\Gamma}\mathbf{1}=\mathbf{q}(\mathbf{B}),
\;
\mathbf{\Gamma}^{\top}\mathbf{1}=\mathbf{p}
\}
$, and $H(\mathbf{\Gamma})=-\sum_{i,j}\Gamma_{ij}(\log \Gamma_{ij}-1)$ is the entropy regularizer. The construction of PairAlign involves three key components, namely the added-budget distribution $\mathbf{q}(\mathbf{B})$, the shortage distribution $\mathbf{p}$, and the transport geometry $\mathbf{C}$.

Here, $\mathbf{q}(\mathbf{B})$ denotes the distribution over candidate added edges induced by the current edge-addition variables $\mathbf{B}$. It represents how the finite added structural budget is placed over candidate non-edges before alignment and is used as the source marginal in the OT problem.

To make the shortage target more concentrated, PairAlign starts from the pairwise shortage of the original graph and derives a target distribution $\mathbf{p}$ by centering it against the graph-wide baseline. This centering gives more weight to severe communication shortages. Specifically, the base shortage $\{\mathcal{S}_{uv}\}_{(u,v)\in\mathcal{P}}$ on the original graph is obtained from Eq.~\eqref{eq:shortage definition} with $\mathbf{W}=\mathbf{A}$. This pattern contains both graph-wide background shortage and genuinely prominent shortage mass. Directly using the raw shortage pattern would spread the budget over many mildly deficient pairs and make the severe regions less visible. We use a simple construction adapted to the graph in which the original shortage is centered relative to a graph-level baseline and only the positive excess $\tilde{p}_{uv}=\max(\mathcal{S}_{uv}-|\mathcal{P}|^{-1}\sum_{(a,b)\in\mathcal{P}}\mathcal{S}_{ab},\,0)$ is retained.
The corresponding shortage target distribution is then

defined by
\begin{equation}
\label{eq:normlized distribution}
\mathbf{p}=(p_{uv})_{(u,v)\in\mathcal P},
\quad
\operatorname{where}
\quad
p_{uv}
=
\frac{\tilde{p}_{uv}}{\sum_{(a,b)\in\mathcal P}\tilde{p}_{ab}}.
\end{equation}

In PairAlign, the transport cost matrix $\mathbf{C}$ captures the structural match between candidate added edges and target shortage pairs. For any candidate added edge $e=(a,b)$ and target pair $(u,v)$, the cost $C_{e,(u,v)}$ has two parts. Endpoint alignment measures whether the edge can connect the source side of the target pair to its target side, while bridge suitability measures whether the span of the candidate edge is commensurate with the scale of the shortage pair. Based on this design, the transport cost is defined by
\begin{align}
C_{(a,b),(u,v)}
={}&
\underbrace{
\begin{aligned}[t]
\min \bigl\{\, d(a,u)+d(b,v),
               d(a,v)+d(b,u)\,\bigr\}
\end{aligned}
}_{{\text{endpoint alignment}}}
\notag
\\
&\quad
-\underbrace{
\lambda
\psi\!\left(
\frac{d(a,b)}{d(u,v)+\varepsilon}
\right)
}_{{\text{bridge suitability}}},
\label{eq:transport_cost}
\end{align}
where $\lambda>0$ controls the strength of the bridge reward and
$\psi(\cdot)$ is a monotone increasing function. The endpoint-alignment term
measures how well a feasible candidate edge is positioned with respect to the
two sides of a target pair. For a single target pair $(u,v)$, the direct shortcut
$(u,v)$, if feasible, is naturally one of the most strongly aligned candidates.
PairAlign does not exclude this case. At the graph level, the goal is not to
add one direct shortcut for each pair with high shortage, but to select a finite set
of added edges that can support many shortage targets under a limited budget.
The transport cost is used to evaluate the structural fit between each feasible
candidate edge and each shortage target. The bridge-suitability term complements the endpoint score by considering the
scale of the candidate edge. Among candidates with similar endpoint alignment,
an edge whose span covers a larger normalized portion of the original pairwise
distance receives a lower cost. In this way, the matrix $\mathbf C$ provides a
support-aware geometry for OT allocation. It favors candidate edges that
are well positioned for a shortage pair and have a span suitable for supporting
that pair under the finite edge budget. If each unit of candidate-edge budget is assigned independently to its cheapest target, the induced target marginal
follows the cost-induced preference instead of the shortage distribution $\mathbf p$. Part of the high-shortage target set can then remain
uncovered. Proposition~\ref{prop:v5_ot_coverage} formalizes the
coverage advantage of shortage alignment. OT alignment addresses it by solving a minimum-cost global
allocation with prescribed added edge and shortage target marginals.
By keeping the allocation aligned with pairs with high shortage, the selected
edges provide more effective structural repair under the same budget.
\begin{proposition}[Coverage advantage over greedy target assignment]
\label{prop:v5_ot_coverage}
Let $\mathcal E=\{e_1,\ldots,e_M\}$ be the candidate-edge set and
$\mathcal T=\{t_1,\ldots,t_m\}$ be the set of shortage target pairs; write
$t\in\mathcal T$ for one target pair. Let
$\mathbf p\in\Delta(\mathcal T)$ be the shortage-target distribution,
$\mathbf q(\mathbf B)\in\Delta(\mathcal E)$ be the added-budget distribution.
Let $E_1,\ldots,E_n$ be drawn independently from $\mathbf q(\mathbf B)$.

For each $e\in\mathcal E$, define the greedy target map
$g_C:\mathcal E\to\mathcal T$ and its induced target distribution by
\begin{equation*}
g_C(e)=\arg\min_{t\in\mathcal T}C_{e,t},
\qquad
\pi_C^{\mathbf B}(t)=
\sum_{e\in\mathcal E}q_e(\mathbf B)\mathbf 1\{g_C(e)=t\}.
\end{equation*}
Let $\mathbf{\Gamma}^g\in\mathbb R_+^{n\times m}$ be the empirical greedy coupling over the
realized edge budgets, where
$\mathbf{\Gamma}^g_{i,t}=n^{-1}\mathbf 1\{g_C(E_i)=t\}$. Let $r_{\mathbf{\Gamma}^g}$ be its
target marginal. Define shortage-target coverage by
\begin{equation*}
\mathrm{Cov}_{\mathbf p}(\Gamma)
=
\sum_{t\in\mathcal T}\min\{r_\Gamma(t),p_t\}.
\end{equation*}
Let $\mathbf{\Gamma}^{\rm OT}$ be any OT coupling over the same realized budget
units with target marginal $\mathbf p$, namely
$\Gamma^{\rm OT}\in\Pi_n(\mathbf p)$, where
\begin{equation*}
\Pi_n(\mathbf p)
=
\{\mathbf\Gamma\ge0:\mathbf\Gamma\mathbf 1=n^{-1}\mathbf 1,\ 
\mathbf\Gamma^\top\mathbf 1=\mathbf p\}.
\end{equation*}
Then
$\mathrm{Cov}_{\mathbf p}(\Gamma^{\rm OT})=1$. Moreover, for any
$\delta\in(0,1)$, with probability at least $1-\delta$,
\begin{equation*}
\mathrm{Cov}_{\mathbf p}(\mathbf\Gamma^{\rm OT})-\mathrm{Cov}_{\mathbf p}(\mathbf\Gamma^g)
\ge
\mathrm{TV}(\pi_C^{\mathbf B},\mathbf p)
-
\frac{|\mathcal T|}{2}
\sqrt{\frac{\log(2|\mathcal T|/\delta)}{2n}} .
\end{equation*}
\end{proposition}

Proposition~\ref{prop:v5_ot_coverage} explains the necessity of OT-guided
allocation. Under the randomized finite-sample view, Greedy-Local maps each sampled
candidate edge to its lowest cost target, so the induced target distribution
$\pi_C^{\mathbf B}$ may drift away from the $\mathbf p$ and
leave some shortage mass uncovered. Proposition~\ref{prop:v5_ot_coverage} shows that, with high probability, the
coverage gain of OT over Greedy-Local is controlled by
$\mathrm{TV}(\pi_C^{\mathbf B},\mathbf p)$ up to a finite-sample error term. OT introduces coordination across
edge budgets by matching the shortage marginal, preventing the
finite budget from collapsing onto a few targets. This coordinated allocation
helps PairAlign turn limited edge additions into targeted shortage reduction and
alleviate over-squashing more effectively.

\textbf{Total Rewiring Objective of PairAlign.} The goal of rewiring in PairAlign is to reduce pairwise communication
shortage under a limited structural budget. OT-guided allocation
controls how structural additions are deployed. The overall objective
combines two terms. The first term measures the focused pairwise
shortage on the rewired graph. Following
Proposition~\ref{prop:bridge_service_displacement}, this term checks
whether the selected insertions increase finite-hop support after row
normalization, and encourages rewiring to reduce the residual
demand--support mismatch of high-shortage pairs,
\begin{equation}
\label{eq:pair-shortage}
L_{\mathrm{shortage}}(\mathbf{W})
=
\sum_{(u,v)\in\mathcal{P}}
 p_{uv}\,\mathcal{S}_{uv}(\mathbf{W}).
\end{equation}

The second term is the OT-guided rewiring term introduced above. The optimal coupling $\mathbf{\Gamma}^{\star}$ represents how candidate added edges should be matched to shortage pairs. Minimizing the OT-guided alignment term yields an allocation with global structural compatibility, defined by
\begin{equation}
\label{eq:OTloss}
L_{\mathrm{OT}}(\mathbf{B})
=
\min_{\Gamma \in \Gamma(\mathbf{q}_{\mathrm{add}}(\mathbf{B}),\mathbf{p})}
\;
\langle \Gamma,\mathbf{C}\rangle
-
\varepsilon H(\Gamma).
\end{equation}

In all, the total objective of PairAlign is defined by combining shortage reduction with OT-guided alignment,
\begin{equation}
\label{eq:main_objective}
\min_{\mathbf{B}}
\quad
L_{\mathrm{shortage}}(\mathbf{A}+\mathbf{B})
+
\lambda_{\mathrm{OT}}
L_{\mathrm{OT}}(\mathbf{B}).
\end{equation}



\subsection{Optimization and Computational Complexity}
\begin{algorithm}[t]
\caption{The PairAlign Rewiring}
\label{alg:PairAlign}
\begin{algorithmic}[1]
\REQUIRE Original graph $G=(V,E,\mathbf{H})$, candidate non-edge set $\bar{E}$, add budget $k$, propagation depth $K$, temperature $\tau$, optimization steps $T_{\mathrm{opt}}$
\ENSURE Discrete added-edge set $E_{\mathrm{add}}$

\STATE \textbf{Phase I: Construct pairwise shortage-aware targets.}
\STATE Compute pairwise demand $\omega_{uv}$ for $(u,v)\in\mathcal{P}$;
\STATE Compute original pairwise shortage $\mathcal{S}_{uv}$ on $\mathbf{A}$ and construct target weights $p_{uv}$;
\STATE Initialize candidate logits $\{\theta_e\}_{e\in\bar{E}}$;

\STATE \textbf{Phase II: Optimize budgeted near-discrete selection.}
\FOR{$t=1$ to $T_{\mathrm{opt}}$}
    \STATE Compute soft edge scores $z_e=\operatorname{softmax}(\boldsymbol{\theta}/\tau)_e$;
    \STATE Select the top-$k$ candidates by a hard mask in the forward pass
    \STATE Form $\mathbf{W}^{(t)}=\mathbf{A}+\mathbf{B}^{(t)}$;
    \STATE Evaluate pairwise supports $s(\mathbf{W}^{(t)})$;
    \STATE Compute focused shortage $L_{\mathrm{shortage}}(\mathbf{W}^{(t)})$;
    \STATE Construct $\mathbf q_{\mathrm{add}}(\mathbf{B}^{(t)})$ and evaluate $L_{\mathrm{OT}}(\mathbf{B}^{(t)})$;
    \STATE Update $\boldsymbol{\theta}$ by minimizing Eq.~\eqref{eq:main_objective} using a straight-through style gradient approximation;
\ENDFOR

\STATE \textbf{Phase III: Project to the final discrete rewiring}
\STATE Rank candidate non-edges by the optimized scores;
\STATE Form a projection pool from the top-ranked candidates;
\STATE Select $E_{\mathrm{add}}$ from the pool by re-evaluating Eq.~\eqref{eq:main_objective} under the add budget;
\RETURN $E_{\mathrm{add}}$
\end{algorithmic}
\end{algorithm}
\textbf{Optimization.} The PairAlign rewiring induced in Eq.~\eqref{eq:main_objective} amounts to choosing a finite subset of added edges from the candidate non-edge set $\bar{E}$, which makes the optimization combinatorial. The candidate non-edge set $\bar{E}$ specifies the feasible locations for edge addition, while $\mathbf{B}$ is the decision matrix supported on $\bar{E}$, with each nonzero entry $B_{uv}$ representing the strength or selection of the added edge $(u,v)$. Direct optimization over such discrete edge subsets is intractable in general. To address this, we introduce a continuous parameterization that yields a differentiable approximation while remaining consistent with the final discrete rewiring. In all, the PairAlign procedure in Alg.~\ref{alg:PairAlign} consists of three stages: shortage-aware pairwise target construction, budgeted near-discrete rewiring, and final projection to the discrete added-edge set.

Specifically, Optimization is carried out over continuous logits on candidate non-edges. Let $\boldsymbol{\theta}=\{\theta_e\}_{e\in\bar{E}}$ denote the collection of all logits. The logits are initialized uniformly, e.g., $\theta_e=0$ for all candidate edges. Let
$z_e=\operatorname{softmax}(\boldsymbol{\theta}/\tau)_e$
be the corresponding soft selection probability under temperature $\tau > 0$.
During optimization, under an add-only rewiring budget of $k$ edges, a top-$k$ hard mask is applied in the forward pass so that only the highest-scoring candidate edges contribute to the current rewiring matrix, with reduced forward computation. In the backward pass, gradients are propagated through the corresponding soft variables by a straight-through style approximation~\cite{bengio2013estimating}. This yields a near-discrete training procedure that remains compatible with gradient-based optimization while staying close to the target discrete add-only rewiring problem.
This objective implements the feedback loop in Fig.~\ref{fig:pairalign_mechanism}.
Updates to $\mathbf B$ change the rewired graph $\mathbf W=\mathbf A+\mathbf B$, the shortage term re-evaluates finite-hop support on the current graph, and the OT term keeps the edge budget aligned with the shortage targets.

After optimization, the learned candidate scores are converted into a discrete rewiring set through a projection step. Candidate non-edges are first ranked by their optimized scores, and a projection pool is formed from the top-ranked entries. The final rewiring set is then selected from this pool under the prescribed add budget by re-evaluating the main objective Eq.~\eqref{eq:main_objective} after optimization. In this way, the final output is a discrete added-edge set, while the training and projection stages remain aligned under a common objective.


\textbf{Computational Complexity.} 
The computational cost of PairAlign mainly comes from truncated propagation evaluation, OT-based alignment, and candidate ranking. Let $M$ be the number of focused target pairs. The one-time construction of the shortage target distribution is a preprocessing cost before the optimization loop. The full sorting of candidate non-edges costs $\mathcal{O}(|\bar{E}|\log |\bar{E}|)$ per step. Dense Sinkhorn alignment over candidate edges and target pairs costs $\mathcal{O}(T_{\mathrm{OT}}|\bar{E}|M)$ per step~\cite{cuturi2013sinkhorn,altschuler2017nearlinear}. The overall optimization complexity is
$ \mathcal{O}\!\left(T_{\mathrm{opt}}\bigl(|\bar{E}|\log |\bar{E}| + T_{\mathrm{OT}}|\bar{E}|M\bigr)\right).
$
If the implementation restricts Sinkhorn alignment to the candidates retained by the hard top-$k$ mask, the factor $|\bar{E}|$ in the OT term is replaced by at most $k$. After optimization, the final candidate scores are ranked once more, adding another $\mathcal{O}(|\bar{E}|\log |\bar{E}|)$ cost. The final projection is performed on a small projection pool, so the exact reevaluation of feasible $k$-subsets remains a low-cost post-processing step. PairAlign is used as an offline rewiring preprocessor, so this cost is incurred once before downstream message-passing model training. A broader comparison with representative rewiring methods is provided in the supplementary appendix.


\begin{table*}[t]
\centering
\caption{Node classification accuracy (\%) on six datasets under different rewiring methods and backbones.The average ranking (AR) reflects the mean position of each method across all
datasets.}
\label{tab:rewiring_nodecls_transposed}
\begin{tabular}{llccccccc}
\toprule
\textbf{Backbone} & \textbf{Rewiring} 
& \textbf{Cora} 
& \textbf{Citeseer} 
& \textbf{Texas} 
& \textbf{Cornell} 
& \textbf{Wisconsin} 
& \textbf{Chameleon} & \textbf{AR}\\
\midrule
\multirow{5}{*}{\textbf{GCN}}
& None~\cite{kipf2016semi} 
& 86.7 $\pm$ 0.3 
& 72.3 $\pm$ 0.3 
& 44.2 $\pm$ 1.5 
& 41.5 $\pm$ 1.8 
& 44.6 $\pm$ 1.4 
& 59.2 $\pm$ 0.6
& 6.9\\
& SDRF~\cite{toppingunderstanding} 
& 86.3 $\pm$ 0.3 
& 72.6 $\pm$ 0.3 
& 43.9 $\pm$ 1.6 
& 42.2 $\pm$ 1.5 
& 46.2 $\pm$ 1.2 
& 59.4 $\pm$ 0.5 
& 5.9 \\
& FoSR~\cite{karhadkar2023fosr} 
& 85.9 $\pm$ 0.3 
& 72.3 $\pm$ 0.3 
& 46.0 $\pm$ 1.6 
& 40.2 $\pm$ 1.6 
& 48.3 $\pm$ 1.3 
& 59.3 $\pm$ 0.6 
& 6.1\\
& BORF~\cite{nguyen2023revisiting} 
& \underline{87.5 $\pm$ 0.2}
& \underline{73.8 $\pm$ 0.2}
& 49.4 $\pm$ 1.2 
& 50.8 $\pm$ 1.1 
& 50.3 $\pm$ 0.9 
& 61.5 $\pm$ 0.4 
& 2.8\\
& GTR~\cite{black2023understanding}
& 87.3 $\pm$ 0.4
& 72.4 $\pm$ 0.3
& 45.9 $\pm$ 1.9
& 50.8 $\pm$ 1.6
& 46.7 $\pm$ 1.5
& 57.6 $\pm$ 0.8
& 5.0 \\
& LASER~\cite{barberolocality}
& 86.9 $\pm$ 1.1 
& 72.6 $\pm$ 0.6
& 45.9 $\pm$ 2.6 
& 42.7 $\pm$ 2.6
& 46.0 $\pm$ 2.6
& 53.7 $\pm$ 1.4 
& 5.7\\
& GOKU~\cite{liangmitigating}
& 86.8 $\pm$ 0.3
& 73.6 $\pm$ 0.2
& \textbf{64.7 $\pm$ 2.4}
& \textbf{58.9 $\pm$ 1.6}
& \textbf{59.8 $\pm$ 2.4}
& \underline{63.2 $\pm$ 0.4} 
& \underline{2.2} \\
\rowcolor{mygray}  & PAR (Ours) 
& \textbf{89.1 $\pm$ 0.4} 
& \textbf{76.8 $\pm$ 0.4} 
& \underline{58.7 $\pm$ 1.6}
& \underline{51.7 $\pm$ 2.8}
& \underline{52.4 $\pm$ 1.7}
& \textbf{65.1 $\pm$ 0.6} 
& \textbf{1.5}\\
\midrule
\multirow{5}{*}{\textbf{GIN}}
& None~\cite{xu2018powerful} 
& 76.0 $\pm$ 0.6 
& 59.3 $\pm$ 0.9 
& 53.5 $\pm$ 3.1 
& 36.5 $\pm$ 2.2 
& 48.5 $\pm$ 2.2 
& 58.1 $\pm$ 2.1 
& 6.1 \\
& SDRF~\cite{toppingunderstanding} 
& 74.9 $\pm$ 0.1 
& 60.3 $\pm$ 0.8 
& 50.3 $\pm$ 3.7 
& 40.0 $\pm$ 2.1 
& 48.8 $\pm$ 1.9 
& 58.4 $\pm$ 2.1 
& 5.7 \\
& FoSR~\cite{karhadkar2023fosr} 
& 75.1 $\pm$ 0.8 
& 61.7 $\pm$ 0.7 
& 47.0 $\pm$ 3.7 
& 35.6 $\pm$ 2.4 
& 48.5 $\pm$ 2.1 
& 56.3 $\pm$ 2.2 
& 6.9\\
& BORF~\cite{nguyen2023revisiting} 
& 78.4 $\pm$ 0.4 
& 63.1 $\pm$ 0.8 
&\underline{63.1 $\pm$ 1.7} 
& 48.6 $\pm$ 1.2 
& 54.9 $\pm$ 1.2 
& \underline{65.3 $\pm$ 0.8} 
& 3.1\\
& GTR~\cite{black2023understanding}
& 78.6 $\pm$ 1.3
& 62.6 $\pm$ 1.9
& 49.5 $\pm$ 2.9
& 39.4 $\pm$ 2.3
& 44.2 $\pm$ 2.4
& 57.1 $\pm$ 1.2
& 5.8\\
& LASER~\cite{barberolocality}
& \underline{79.1 $\pm$ 1.0} 
&\textbf{66.5 $\pm$ 1.3}
& 46.5 $\pm$ 4.5 
& 44.5 $\pm$ 3.8
& 46.1 $\pm$ 4.6
& 59.8 $\pm$ 2.2 
& 4.5\\
& GOKU~\cite{liangmitigating}
& 78.4 $\pm$ 0.5
& 63.6 $\pm$ 1.3
& 60.2 $\pm$ 2.6
& \underline{49.5 $\pm$ 3.5}
& \underline{57.6 $\pm$ 3.1}
& 62.1 $\pm$ 0.6 
& \underline{2.9}\\
\rowcolor{mygray} & PAR (Ours) 
& \textbf{80.6 $\pm$ 0.6} 
& \underline{65.9 $\pm$ 1.1} 
& \textbf{68.8 $\pm$ 2.8} 
& \textbf{51.0 $\pm$ 2.9} 
& \textbf{58.2 $\pm$ 1.9} 
& \textbf{65.5 $\pm$ 0.7} 
& \textbf{1.0}\\
\bottomrule
\end{tabular}
\end{table*}
\section{Experiments}
\subsection{Experimental Setup}
The experimental study compares PairAlign (PAR) with representative preprocessing-based graph rewiring baselines covering the main design families in the literature. Specifically, we include curvature-driven methods, namely SDRF~\cite{toppingunderstanding} and BORF~\cite{nguyen2023revisiting}; spectral or global-connectivity methods, including FoSR~\cite{karhadkar2023fosr}, the effective resistance-based rewiring method GTR~\cite{black2023understanding}, and GOKU~\cite{liangmitigating}; and the locality-aware sparse rewiring method LASER~\cite{barberolocality}. For the heterophilic benchmarks, we further compare with two recent variants that extend rewiring beyond purely structural criteria, namely JDR~\cite{linkerhagnerjoint}, which couples graph rewiring with feature denoising, and ComFy~\cite{rubiognns}, which guides rewiring using community structure and feature similarity. For each backbone, we also report the corresponding model on the original graph as the no-rewiring baseline.
Across all tables, boldface denotes the best result and underline denotes the second-best result.



The benchmarks are organized into three groups. The first group contains widely used real-world node classification datasets: the citation networks Cora and CiteSeer~\cite{yang2016revisiting}, the web graphs Texas, Cornell, and Wisconsin~\cite{pei2020geom}, and the Wikipedia-network dataset Chameleon~\cite{rozemberczki2021multi}. The second group contains the standard TUDataset graph classification benchmarks ENZYMES, IMDB-BINARY, MUTAG, PROTEINS, REDDIT-BINARY, and COLLAB~\cite{morris2020tudataset}. The third group contains five heterophilic node classification benchmarks from Platonov et al.~\cite{platonov2023critical}: Roman-Empire, Amazon-Ratings, Minesweeper, Tolokers, and Questions. Following the standard evaluation protocols of these benchmarks, performance is reported in classification accuracy for the first group and for Roman-Empire and Amazon-Ratings, and in ROC AUC for Minesweeper, Tolokers, and Questions. Detailed dataset statistics, split protocols, and implementation settings are provided in the appendix.

For the standard node classification and graph classification benchmarks, we consider two widely used message-passing backbones, namely GCN~\cite{kipf2016semi} and GIN~\cite{xu2018powerful}. For the heterophilic node classification benchmarks, we use GCN~\cite{kipf2016semi} and GAT~\cite{velickovic2018graph}. Within each benchmark family, all rewiring methods are evaluated under the same backbone, data split, and training protocol unless otherwise stated, so that performance differences can be attributed to the rewiring strategy rather than to backbone-specific tuning. Rewiring is treated as a structural preprocessing step, after which the downstream message-passing model is trained on the rewired graph using an otherwise identical optimization pipeline across competing methods. This protocol is designed to isolate the contribution of rewiring to both graph structure and downstream prediction.

\subsection{Evaluation Protocol}
We evaluate rewiring from two complementary perspectives, downstream prediction performance and structural rewiring quality. For downstream prediction performance, the primary metric follows the native evaluation protocol of each benchmark family. Classification accuracy is reported on node classification and graph classification datasets. We state the exact evaluation implementation in Appendix and report results under that protocol for reproducibility.

Beyond downstream accuracy, we evaluate how rewiring changes the communication structure of the graph. We report two pair-centric diagnostics, $\Delta$Shortage and Coverage@10, to inspect whether the added edges repair pairs with high shortage. Effective resistance proxies provide external references for pairwise and global communication bottlenecks.

Based on the pair-shortage $\mathcal{S}_{uv}$ in Eq.~\eqref{eq:shortage definition} and the normalized shortage weight $p_{uv}$ in Eq.~\eqref{eq:normlized distribution} on the computational graph $\mathbf W$, we measure the relative reduction of shortage on the rewired graph $\mathbf W'$ by
\begin{equation*}
\Delta\mathrm{Shortage}
=
\frac{
\sum_{(u,v)\in\mathcal{T}}
 p_{uv}
\left[
\mathcal{S}_{uv}(\mathbf W)-\mathcal{S}_{uv}(\mathbf W')
\right]
}{
\sum_{(u,v)\in\mathcal{T}}
 p_{uv}\mathcal{S}_{uv}(\mathbf W)+\varepsilon
}.
\end{equation*}
A larger value indicates stronger repair of the originally underserved pairs. Coverage@10 measures how broadly this repair reaches the most severe shortage region. Let $\mathcal{T}_{10}\subset\mathcal{T}$ be the top 10\% pairs ranked by $\mathcal{S}_{uv}(\mathbf W)$. We compute
\begin{equation*}
\mathrm{Coverage@10}
=
\frac{1}{|\mathcal{T}_{10}|}
\sum_{(u,v)\in\mathcal{T}_{10}}
\mathbb{I}
\left[
s_{uv}(\mathbf W')-s_{uv}(\mathbf W)>\eta
\right],
\end{equation*}
where $\eta$ is a small numerical tolerance. The value $\Delta$Shortage measures the amount of targeted repair, and Coverage@10 measures the fraction of top shortage pairs that receive nontrivial support. 


We further report effective resistance proxies to relate our diagnostics to established bottleneck measures~\cite{black2023understanding}. For a pair $(u,v)$, the pairwise effective resistance is
\begin{equation*}
\mathrm{PER}_{G}(u,v)
=
(\mathbf{e}_u-\mathbf{e}_v)^{\top}
L_G^{\dagger}
(\mathbf{e}_u-\mathbf{e}_v),
\end{equation*}
where $L_G^{\dagger}$ is the Moore--Penrose pseudoinverse of the graph Laplacian, and $\mathbf e_u,\mathbf e_v\in\mathbb R^{|V|}$ are the standard basis vectors associated with nodes $u$ and $v$, respectively. At the graph level, we use total effective resistance,
\begin{equation*}
\mathrm{TER}(G)
=
\sum_{u<v}
\mathrm{PER}_{G}(u,v)
=
|V|
\sum_{i=2}^{|V|}
\frac{1}{\lambda_i(L_G)}.
\end{equation*}
Here, $\lambda_i(L_G)$ is the $i$-th eigenvalue of $L_G$. Lower PER or TER indicates smaller effective resistance and stronger communication connectivity. We report the relative reductions of PER on top shortage pairs and TER on the whole graph as external bottleneck proxies. Rewiring is applied as a preprocessing step, and downstream results are averaged over repeated seeds.

\subsection{Experiment analysis}
\begin{table*}[t]
\centering
\caption{Graph classification accuracy (\%) on six datasets under different rewiring methods and backbones. The average ranking (AR) reflects the mean position of each method across all
datasets.}
\label{tab:graph_classification_final}
\begin{tabular}{llccccccc}
\toprule
\textbf{Backbone} & \textbf{Rewiring} & \textbf{ENZYMES} & \textbf{IMDB} & \textbf{MUTAG} & \textbf{PROTEINS} & \textbf{REDDIT} & \textbf{COLLAB} & \textbf{AR}\\
\midrule
\multirow{5}{*}{\textbf{GCN}}
& None~\cite{kipf2016semi}     & 25.5 $\pm$ 1.3 & 49.3 $\pm$ 1.0 & 68.8 $\pm$ 2.1 & 70.6 $\pm$ 1.0 & 81.4 $\pm$ 1.7 & 67.2 $\pm$ 1.2 & 7.1\\
& SDRF~\cite{toppingunderstanding}     & 26.1 $\pm$ 1.1 & 49.1 $\pm$ 0.9 & 70.5 $\pm$ 2.1 & 71.4 $\pm$ 0.8 & 85.4 $\pm$ 1.4 & 67.2 $\pm$ 1.4 & 6.0\\
& FoSR~\cite{karhadkar2023fosr}     & 27.4 $\pm$ 1.1 & 49.6 $\pm$ 0.8 & 75.6 $\pm$ 1.7 & \underline{72.3 $\pm$ 0.9} & \underline{86.2 $\pm$ 0.9} & 65.3 $\pm$ 1.2 & 4.4\\
& BORF~\cite{nguyen2023revisiting}     & 24.7 $\pm$ 1.0 & \underline{50.1 $\pm$ 0.9} & 75.8 $\pm$ 1.9 & 71.0 $\pm$ 0.8 & 85.9 $\pm$ 2.1 & TIMEOUT & 5.3\\
& GTR~\cite{black2023understanding}  & 27.4 $\pm$ 1.1 & 49.5 $\pm$ 1.0 & 78.9 $\pm$ 1.8 & 72.4 $\pm$ 1.2 & 84.8 $\pm$ 1.1  & 67.0 $\pm$ 1.0 & 4.7\\
& LASER~\cite{barberolocality} &\underline{27.7} $\pm$ 2.6 & 49.9 $\pm$ 1.6 & 75.2 $\pm$ 3.6 & \textbf{72.7 $\pm$ 1.7} & 84.8 $\pm$ 1.9 & \underline{67.6 $\pm$ 1.2} & \underline{3.4}\\
& GOKU~\cite{liangmitigating} & 27.6 $\pm$ 1.2 & 49.8 $\pm$ 0.7 & \underline{81.0 $\pm$ 2.0} & 71.9 $\pm$ 0.8 & 85.4 $\pm$ 1.2 & 67.5 $\pm$ 1.0 & 3.5\\
\rowcolor{mygray}  & PAR (Ours) & \textbf{31.5 $\pm$ 3.8} & \textbf{50.3 $\pm$ 2.5} & \textbf{83.3 $\pm$ 0.8} & 71.9 $\pm$ 1.6  & \textbf{88.9 $\pm$ 1.7} & \textbf{68.2 $\pm$ 1.7} & \textbf{1.6}\\
\midrule
\multirow{5}{*}{\textbf{GIN}}
& None~\cite{xu2018powerful}     & 31.3 $\pm$ 1.2 & 69.0 $\pm$ 1.3 & 75.5 $\pm$ 2.9 & 69.7 $\pm$ 1.0 & 87.6 $\pm$ 3.5 & 73.1 $\pm$ 0.7 & 6.8\\
& SDRF~\cite{toppingunderstanding}     & 33.5 $\pm$ 1.3 & 68.6 $\pm$ 1.2 & 77.3 $\pm$ 2.3 & 72.2 $\pm$ 0.9 & 89.0 $\pm$ 1.1 & 73.6 $\pm$ 0.7 & 4.9\\
& FoSR~\cite{karhadkar2023fosr}     & 25.3 $\pm$ 1.2 & 69.5 $\pm$ 1.1 & 75.2 $\pm$ 3.0 & \textbf{74.2 $\pm$ 0.8} & \underline{89.4 $\pm$ 0.9} & \underline{74.0 $\pm$ 0.9} & 4.3\\
& BORF~\cite{nguyen2023revisiting}     & \underline{35.5} $\pm$ 1.2 & \underline{71.3 $\pm$ 1.5} &\underline{80.8 $\pm$ 2.5} & 71.3 $\pm$ 1.0 & \textbf{89.9 $\pm$ 0.9} & TIMEOUT & 3.8\\
& GTR~\cite{black2023understanding}  & 28.4 $\pm$ 1.8 & 70.1 $\pm$ 1.2 & 78.5 $\pm$ 3.5 & 73.3 $\pm$ 0.9 &  88.3 $\pm$ 0.9 & 73.3 $\pm$ 0.8 & 4.7\\
& LASER~\cite{barberolocality} &35.3 $\pm$ 1.3 & 68.6 $\pm$ 1.2 & 76.1 $\pm$ 2.4 & 72.1 $\pm$ 0.7 &  87.1 $\pm$ 0.6&  73.1 $\pm$ 0.9 & 6.0\\
& GOKU~\cite{liangmitigating} & 33.8 $\pm$ 1.2 & \underline{71.3 $\pm$ 0.9} & 78.4 $\pm$ 2.5 & \underline{73.9 $\pm$ 1.0} & 88.8 $\pm$ 0.9 & 73.7 $\pm$ 0.8 & \underline{3.4}\\
\rowcolor{mygray}  & PAR (Ours) & \textbf{44.5 $\pm$ 2.6} & \textbf{71.5 $\pm$ 1.7} &  \textbf{82.6 $\pm$ 3.0}   &  71.7 $\pm$ 1.8              & 89.2 $\pm$ 1.5 & \textbf{74.4 $\pm$ 1.9} & \textbf{2.2}\\
\bottomrule
\end{tabular}
\end{table*}
\textbf{Node Classification}. PAR gives the strongest node classification results under both backbones, with the best average rank under GCN and GIN, reaching 1.50 and 1.00. The gains are steady on citation graphs and become larger on topology-limited datasets. Under GIN, Texas rises from 53.5 to 68.8 and Cornell from 36.5 to 51.0; under GCN, PAR also improves the no-rewiring baseline on both Cora and CiteSeer. The consistent gains across GCN and GIN suggest that the benefit comes from the repaired topology used for message passing. PAR improves the pairwise support available before backbone training, which is especially useful when useful information must move beyond immediate neighborhoods.

\textbf{Graph Classification}. PAR also performs strongly on the TUD graph classification benchmarks, with the best average rank under both GCN and GIN. Under GCN, it attains the top result on five of the six datasets, with clear gains on ENZYMES and MUTAG. Under GIN, it remains strongest on ENZYMES, IMDB, MUTAG, and COLLAB. The gains are larger when graph labels depend on information distributed across separated or weakly connected substructures. They are narrower on PROTEINS and in some REDDIT settings, where local motifs or sufficient original structure may already carry much of the label signal. This pattern supports the intended role of PairAlign. Pairwise support can improve whole-graph representations when labels require information to move across separated parts of the graph.
\begin{table*}[t]
\centering
\caption{Results on heterophilic node classification datasets. Roman-Empire and Amazon-Ratings are reported in Accuracy (\%), while Minesweeper, Tolokers, and Questions are reported in ROC AUC (\%). The average ranking (AR) reflects the mean position of each method across all
datasets.}
\label{tab:hetero_main}
\begin{tabular}{llcccccc}
\toprule
\textbf{Backbone} & \textbf{Rewiring}
& \textbf{Roman-Empire}
& \textbf{Amazon-Ratings}
& \textbf{Minesweeper}
& \textbf{Tolokers}
& \textbf{Questions}
& \textbf{AR} \\
\midrule
\multirow{8}{*}{\textbf{GCN}}
& None~\cite{kipf2016semi}     & 72.6 $\pm$ 0.6 & 47.9 $\pm$ 0.5 & 89.2 $\pm$ 0.7 & 83.2 $\pm$ 1.0 & 76.0 $\pm$ 1.1 & 7.3 \\
& SDRF~\cite{toppingunderstanding}     &     73.7 $\pm$ 0.6             &    48.1 $\pm$ 0.5              &    89.3 $\pm$ 0.4              &       83.5 $\pm$ 0.6           &  76.0 $\pm$ 1.0                &    5.7\\
& FoSR~\cite{karhadkar2023fosr}     &  73.7 $\pm$ 0.7            &   48.6 $\pm$ 0.5               &      \textbf{89.5 $\pm$ 0.4}            &     83.6 $\pm$ 0.6             &   76.3 $\pm$ 0.9               &   3.4\\
& BORF~\cite{nguyen2023revisiting}     &       73.8 $\pm$ 0.6           &         48.8 $\pm$ 0.5         &     89.3 $\pm$ 0.6             &     TIMEOUT             &            TIMEOUT     &   5.2 \\
& GTR~\cite{black2023understanding}      &     73.2   $\pm$ 0.6           & 48.2 $\pm$ 0.4                 &    89.3 $\pm$ 0.5              &    83.6 $\pm$ 0.8              &  76.2 $\pm$ 1.0                &  4.9  \\
& JDR~\cite{linkerhagnerjoint}    &    72.9 $\pm$ 0.7              &    48.2 $\pm$ 0.5              &   89.1 $\pm$ 0.5               &   83.8 $\pm$ 0.8               & \textbf{76.6 $\pm$ 0.9}                 &  4.9  \\
& ComFy~\cite{rubiognns}     &   \underline{73.8 $\pm$ 0.5}               &   \underline{49.2 $\pm$ 0.4}              &   89.3 $\pm$ 0.5               &      \textbf{85.1 $\pm$ 0.7}            & \underline{76.4 $\pm$ 0.9}                 &  \underline{2.4}  \\
 \rowcolor{mygray}  & PAR (Ours) &         \textbf{74.1 $\pm$ 0.7}         &       \textbf{49.6 $\pm$ 0.3}           &         \underline{89.4 $\pm$ 0.6}         &     \underline{84.7 $\pm$ 0.7}            &          76.1 $\pm$ 1.3        &   \textbf{2.2} \\
\midrule
\multirow{8}{*}{\textbf{GAT}}
& None~\cite{velickovic2018graph}     & 80.3 $\pm$ 0.7 & 47.8 $\pm$ 0.7 & 91.3 $\pm$ 0.6 & 81.9 $\pm$ 0.7 & 77.1 $\pm$ 1.0 & 7.3 \\
& SDRF~\cite{toppingunderstanding}     &     80.8 $\pm$ 0.4             &    48.3 $\pm$ 0.5              &     91.3 $\pm$ 0.4             &      82.5 $\pm$ 0.5            &         77.4 $\pm$ 0.8         &  5.3  \\
& FoSR~\cite{karhadkar2023fosr}     &   80.9 $\pm$ 0.4               &    49.0 $\pm$ 0.5              &       91.5 $\pm$ 0.6           &         82.6 $\pm$ 0.5         &           77.4 $\pm$ 0.8       &   3.6 \\
& BORF~\cite{nguyen2023revisiting}     &    80.8 $\pm$ 0.6              & \textbf{49.3 $\pm$ 0.4}                 &     91.1 $\pm$ 0.4             &   TIMEOUT               &           TIMEOUT       &  6.0  \\
& GTR~\cite{black2023understanding}      &   80.7 $\pm$ 0.5               & 48.4 $\pm$ 0.5                 &  91.7 $\pm$ 0.4                &   82.1 $\pm$ 0.6               &          77.3 $\pm$ 0.8        &  5.6  \\
& JDR~\cite{linkerhagnerjoint}    &      80.8 $\pm$ 0.5            &   48.5 $\pm$ 0.5               & 91.7 $\pm$ 0.6                 &       82.1 $\pm$ 0.7           & 77.4 $\pm$ 0.8                  &  4.6  \\
& ComFy~\cite{rubiognns}     &     \textbf{81.1 $\pm$ 0.5}             &          48.8 $\pm$ 0.7        &      \textbf{92.3 $\pm$ 0.5}            &          \underline{82.7 $\pm$ 0.6}        &  \underline{77.5 $\pm$ 0.8}               &  \underline{2.0}  \\
\rowcolor{mygray}  & PAR (Ours) &   \underline{81.0 $\pm$ 0.4}               &    \underline{49.1 $\pm$ 0.5}              &     \underline{92.1 $\pm$ 0.4}            &        \textbf{83.8 $\pm$ 0.6}          &       \textbf{77.6 $\pm$ 0.7}           &  \textbf{1.6}  \\
\bottomrule
\end{tabular}
\end{table*}

\textbf{Heterophilous Node Classification}. The heterophilous benchmarks are a stricter test of rewiring quality because useful evidence is often nonlocal, but arbitrary shortcuts can damage class-relevant structure. PAR obtains the best average rank under both GCN and GAT, improving over the closest competitor ComFy in each case. With GCN, it gives the strongest results on Roman-Empire and Amazon-Ratings and remains close to the top on Minesweeper and Tolokers. With GAT, it leads on Tolokers and Questions and stays within a small margin of the best result on the remaining datasets. This stability is more informative than isolated wins. It shows that PAR can add useful nonlocal support without turning rewiring into broad long-range mixing, which is important for heterophilous graphs where distant evidence and structural specificity must both be preserved.

The benchmark groups offer three important observations about PairAlign.
\begin{itemize}
\item On node classification, PAR improves both GCN and GIN across the
tested datasets and keeps the best average rank under both backbones. This
consistent gain across backbones supports a graph-side explanation. PairAlign
repairs pairwise support for message passing.
\item On graph classification, the larger gains appear on datasets where labels
depend on information spread across separated or weakly connected substructures.
This extends the effect of pairwise repair beyond node prediction, showing that
better pairwise support can improve whole-graph representations.
\item On heterophilous graphs, PAR is more stable across datasets than baselines
that peak only in a few cases. This stability matters because heterophily needs
nonlocal evidence but can be harmed by indiscriminate mixing. PAR adds nonlocal
support through shortage targets, keeping long-range rewiring selective.
\end{itemize}


\subsection{Pair-Shortage Validity.}

\begin{figure}[tbp]
\centering
\subfigure[Pair-shortage validity]{
    \includegraphics[width=0.45\linewidth]{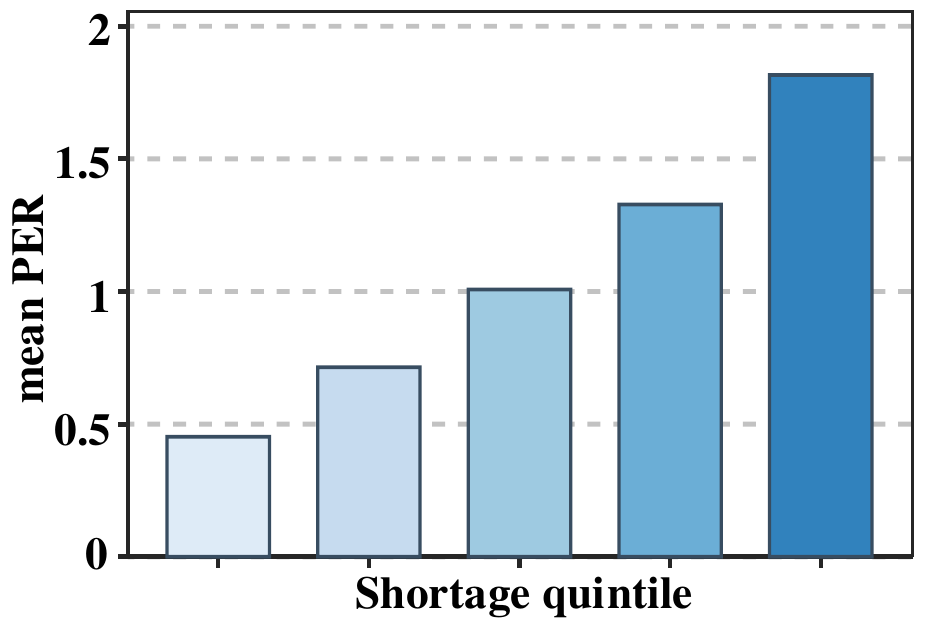}
    \label{fig:per_validity}
}
\hfill
\subfigure[Structural reduction]{
    \includegraphics[width=0.46\linewidth]{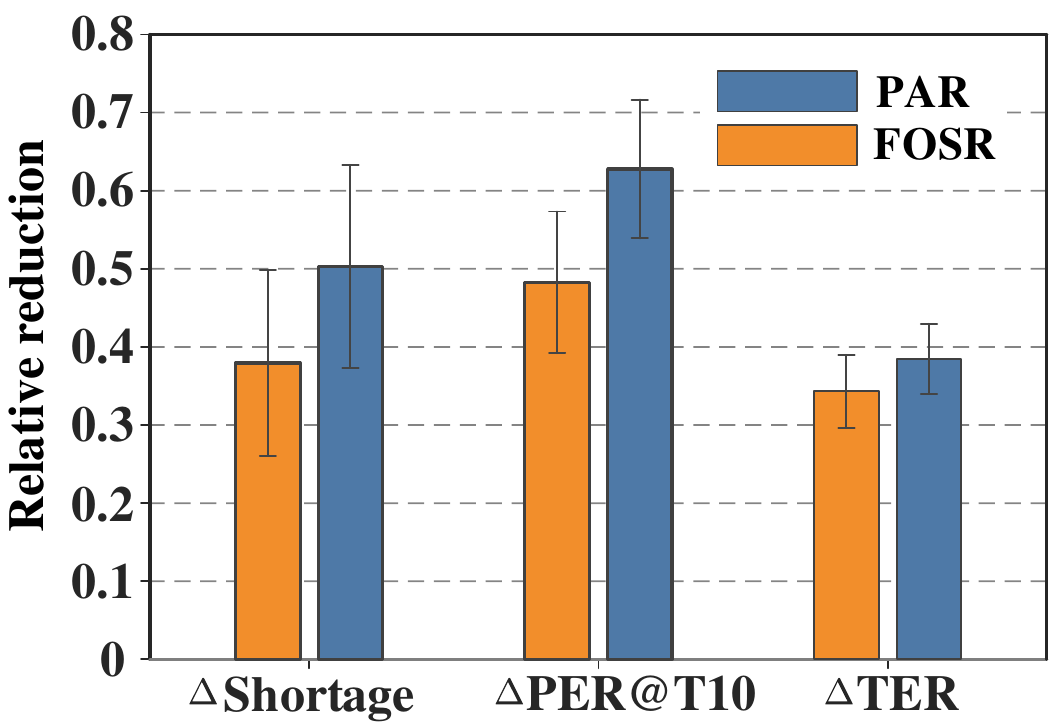}
    \label{fig:structural_reduction}
}
\caption{Pair-level shortage analysis and structural reduction comparison.}
\label{fig:pair_shortage_analysis}
\end{figure}

We first relate the proposed pair-shortage score to an external pair-level bottleneck proxy before analyzing the effect of rewiring. For each graph in MUTAG, we compute the pair-shortage score on the original graph and compare it with pairwise effective resistance (PER). We then group node pairs into five shortage quantiles and report the normalized mean PER in each bin. As shown in Fig.~\ref{fig:pair_shortage_analysis}(a), higher shortage quantiles consistently correspond to larger PER. The normalized mean PER increases from 0.4527 in the lowest shortage bin to 1.8161 in the highest bin, and the graph-level Spearman correlation between shortage and PER reaches 0.9304 on average. This monotone pattern indicates that pairs with high shortage are also pairwise bottlenecks under an effective-resistance view. Hence, the proposed pair-shortage can provide a sensible pair-level diagnostic that is aligned with an established communication bottleneck proxy.

We next evaluate structural reduction on the same target pairs fixed before rewiring. On MUTAG, PAR is compared with FOSR under the same edge budget. Since FOSR directly optimizes a spectral graph-connectivity surrogate, it provides a relevant baseline for resistance-based evaluation.

As shown in Fig.~\ref{fig:pair_shortage_analysis}(b), PAR yields larger reductions in both the shortage score and the pairwise effective resistance of the top 10\% pairs with high shortage. $\Delta\mathrm{PER}@T10$ increases from $0.4826$ with FOSR to $0.6278$ with PAR, and $\Delta$Shortage increases from $0.3795$ to $0.5030$. This indicates that the pair-centric objective of PAR improves aggregate graph connectivity and guides the limited rewiring budget toward communication-deficient node pairs. PAR makes the structural modification more aligned with the pairwise interactions relevant to downstream prediction.
\subsection{Main Ablations.}
The ablation study in Table~\ref{tab:ablation_main} isolates the components that turn shortage diagnosis into targeted repair. We compare the full model with Greedy-Local and variants that remove focused target selection, remove OT allocation, or drop one term in the transport cost. Task performance is reported together with $\Delta$Shortage and Coverage@10.

\begin{table*}[t]
\centering
\caption{Mechanism-oriented ablations on three representative benchmarks. 
Higher values mean better task performance, shortage reduction, and Coverage@10.}
\label{tab:ablation_main}
\begin{tabular}{lccc ccc ccc}
\toprule
\multirow{2}{*}{Method} 
& \multicolumn{3}{c}{Texas} 
& \multicolumn{3}{c}{Roman-Empire} 
& \multicolumn{3}{c}{MUTAG} \\
\cmidrule(lr){2-4} \cmidrule(lr){5-7} \cmidrule(lr){8-10}
& Task & $\Delta$Shortage & Coverage@10
& Task & $\Delta$Shortage & Coverage@10
& Task & $\Delta$Shortage & Coverage@10 \\
\midrule

Greedy-Local   
& \underline{0.5622} & 0.0452 & 0.0351
& 0.7317 & 0.2795 & 0.2143
& 0.7833 & 0.1620 & 0.0467  \\

w/o Focus      
& 0.5568 & \underline{0.2208} & \underline{0.1939}
& 0.7293 & 0.5610 & 0.5857
& 0.7961 & 0.4288 & 0.5479 \\

w/o OT         
& 0.5351 & 0.1741 & 0.1471
& 0.7305 & \underline{0.5709} & 0.6143
& 0.7767 & \underline{0.4826} & \underline{0.6432} \\

w/o Bridge   
& 0.5528 & 0.1586 & 0.1668
& \underline{0.7346} & 0.5487 & \underline{0.6465}
& \underline{0.8078} & 0.4415 & 0.6036 \\

w/o Endpoint   
& 0.5459 & 0.1235 & 0.1036
& 0.7295 & 0.5216 & 0.4952
& 0.8030 & 0.4186 & 0.5368 \\

\midrule
\rowcolor{mygray} PAR       
& \textbf{0.5877} & \textbf{0.2350} & \textbf{0.2113}
& \textbf{0.7412} & \textbf{0.6220} & \textbf{0.7048}
& \textbf{0.8263} & \textbf{0.4855} & \textbf{0.6447} \\
\bottomrule
\end{tabular}
\end{table*}
The full model gives the clearest evidence for this mechanism. PAR obtains the best task performance on Texas, Roman-Empire, and MUTAG, while also producing the largest $\Delta$Shortage and Coverage@10. For example, on Texas, PAR reaches $\Delta$Shortage of $0.2350$ and Coverage@10 of $0.2113$, both higher than all ablated variants. On MUTAG, PAR achieves the best task score of $0.8263$, with the strongest structural diagnostics as well. The consistent improvement in both prediction and structural repair supports the alignment between PAR's downstream gains and its targeted repair of pairs with high shortage.

The focused target set is the first ingredient behind this behavior. PAR does not start from candidate edges alone; it first selects pairs with high shortage on the original graph as the pairs to be repaired by rewiring. Removing the focus mechanism still preserves part of the shortage signal, so the variant does not collapse. Yet \textit{w/o Focus} is consistently below PAR in task performance, $\Delta$Shortage, and Coverage@10 across the three datasets. On Roman-Empire, Coverage@10 decreases from $0.7048$ to $0.5857$, and on MUTAG it drops from $0.6447$ to $0.5479$. This gap means that the focused target set gives rewiring a sharper pairwise target. Greedy-Local further shows why local edge ranking is not enough. It selects edges by local improvement but does not coordinate the budget across shortage targets, so its Coverage@10 is only $0.0351$ on Texas and $0.0467$ on MUTAG, far below PAR. The focus mechanism does more than filter candidates by specifying the pairwise interactions that should receive the limited rewiring budget.

The transport term becomes important once the target set has been fixed. The \textit{w/o OT} variant keeps the focus mechanism, continuous optimization, and projection pipeline, but sets the transport weight to zero. It yields nontrivial $\Delta$Shortage, reaching $0.5709$ on Roman-Empire and $0.4826$ on MUTAG. This confirms that the shortage term can expose communication-deficient regions. The \textit{w/o OT} variant remains below PAR in task performance and targeted coverage. On Texas, Coverage@10 decreases from $0.2113$ to $0.1471$, and on Roman-Empire it decreases from $0.7048$ to $0.6143$. These gaps clarify the role of OT: shortage reveals where support is lacking, and OT coordinates the limited edge budget across targets with high shortage to produce an organized allocation.

The remaining variants separate the roles of the two geometric terms in the transport cost. In \textit{w/o Bridge}, endpoint matching is retained, so the model can still cover a moderate fraction of pairs with high shortage. For instance, on Roman-Empire, its Coverage@10 is $0.6465$, close to \textit{w/o OT} at $0.6143$ and clearly above \textit{w/o Endpoint} at $0.4952$. Its $\Delta$Shortage remains lower than PAR, decreasing from $0.2350$ to $0.1586$ on Texas and from $0.4855$ to $0.4415$ on MUTAG. Endpoint proximity alone does not guarantee effective repair; the bridge reward helps select edges that actually increase pairwise support. In contrast, \textit{w/o Endpoint} keeps the bridge reward but suffers a larger drop in Coverage@10, especially on Texas, where it decreases from $0.2113$ to $0.1036$. Bridge-like edges still need to be aligned with the intended target pairs. The two cost terms play distinct roles. Endpoint matching determines the pairwise targets an added edge should support, and the bridge reward favors edges that can produce effective communication repair.

By reformulating over-squashing around communication-deficient node pairs, PAR recasts graph rewiring as a budgeted structural repair problem, yielding stronger targeted repair and better downstream performance across the datasets.

\subsection{Analysis of OT-guided Budget Allocation}
\begin{figure}
\centering
{\includegraphics[width=1\linewidth]{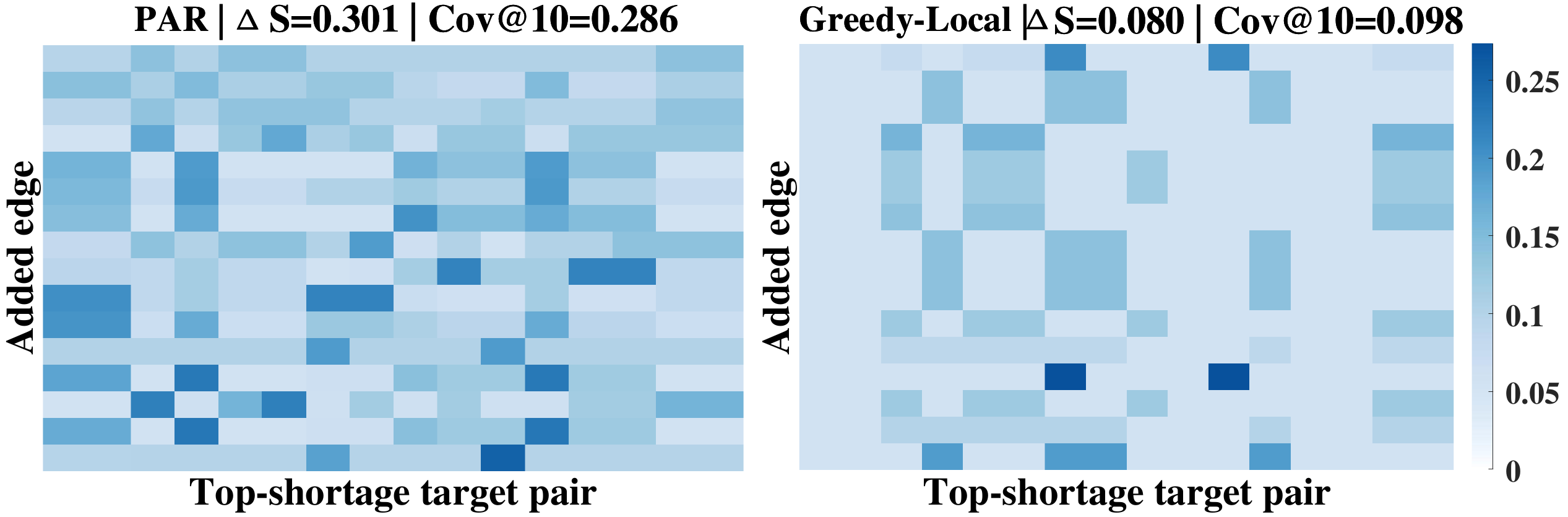}}
\caption{Visualization of how added edges are allocated to target pairs with high shortage. Each row is normalized to show how one added edge distributes its allocation over the selected target pairs.
Darker colors indicate a larger allocated fraction for the corresponding target pair.}
\label{fig: heatmap}
\end{figure}

\textbf{Necessity of OT.} To observe how the OT-guided rewiring rule changes the allocation of a limited edge addition budget, we visualize the assignment from the final added edges to the top shortage target pairs. As shown in Fig.~\ref{fig: heatmap}, each row corresponds to the added edge and each column corresponds to a target pair with high shortage. Darker entries indicate that a larger fraction of the allocation associated with an added edge is assigned to the corresponding target pair.

The heatmap shows that our PAR distributes its added edges across a broader set of targets with high shortage while retaining clear assignments. Full PAR achieves higher $\Delta$Shortage and Coverage@10. Greedy-Local exhibits a more concentrated allocation pattern, with several added edges assigned mainly to a small subset of target columns. These results indicate that OT-guided allocation spreads the budget over a wider target set and produces stronger pair-centric structural repair.

\begin{figure*}[!t]
\centering
\subfigure[OT weight]{
    \includegraphics[width=0.295\textwidth]{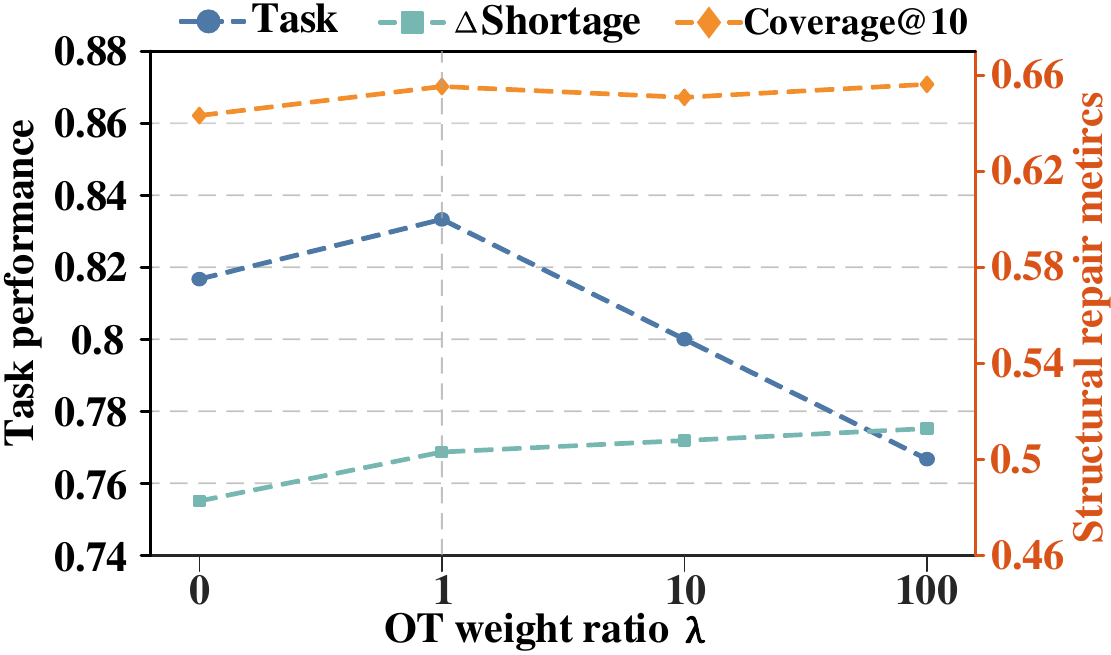}
    \label{fig: stability}
}
\hspace{0.02\textwidth}
\subfigure[Rewiring budget]{
    \includegraphics[width=0.295\textwidth]{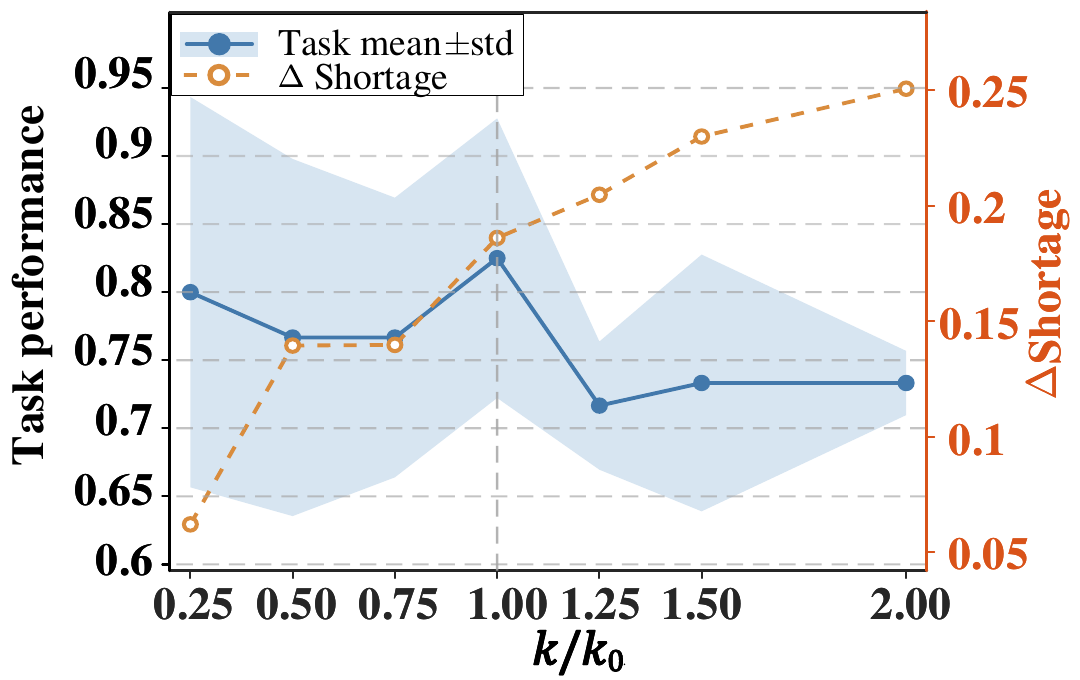}
    \label{fig:budget_sensitivity}
}
\hspace{0.02\textwidth}
\subfigure[Propagation depth]{
    \includegraphics[width=0.295\textwidth]{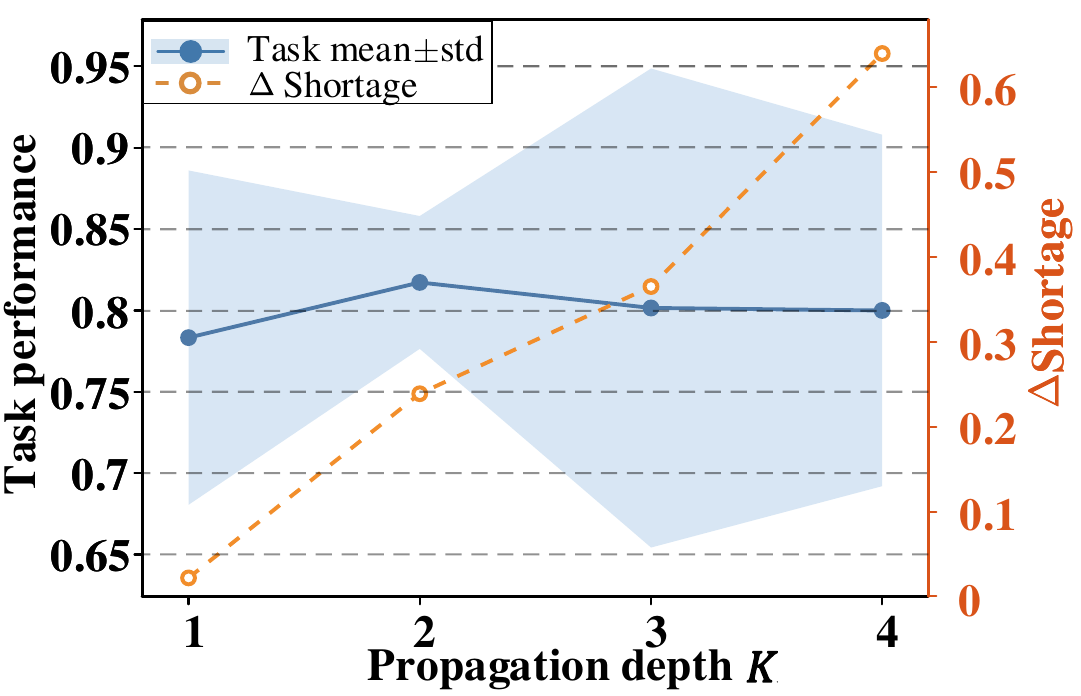}
    \label{fig:capacity}
}
\caption{Visualization of OT-weight stability and sensitivity analysis.}
\label{fig: Sensitivity}
\end{figure*}

\textbf{Stability of OT.}
We further analyze how the weight $\lambda_{\mathrm{OT}}$ of the OT-guided rewiring affects the rewiring objective. The MUTAG-GCN setting is kept fixed and only $\lambda_{\mathrm{OT}}$ is varied. As shown in Fig.~\ref{fig: stability}, OT-guided alignment improves Task and structure proxies in the main operating range compared with $\lambda_{\mathrm{OT}}=0$. This indicates that the OT-guided rule provides a proper global coordination signal for structural repair and helps added edges cover targets with high shortage more effectively. When $\lambda_{\mathrm{OT}}$ is further increased, $\Delta$Shortage remains high and even continues to increase, but Task performance starts to decrease. This suggests that an overly strong transport preference mainly emphasizes the structural repair objective and does not necessarily translate into better prediction. The trend supports the role of OT as a budget-coordination mechanism. With a suitable weight, it organizes the limited edge addition budget across shortage targets and helps convert structural repair into downstream task gains.

\subsection{Sensitivity Analysis.}

\textbf{Rewiring budget.}
The budget sensitivity on MUTAG is shown in Fig.~\ref{fig:budget_sensitivity}. We set $k_0=3$ as the integer PAR budget anchor compatible with the main GIN setting. The task curve shows a high-variance plateau; the highest mean appears at $0.25k_0$, and its error band largely overlaps with the results from $0.5k_0$ to $k_0$. In this compact budget range, PAR already achieves competitive task performance, and $\Delta$Shortage increases from $0.0620$ to $0.1860$. When the budget is enlarged beyond $k_0$, $\Delta$Shortage continues to rise and reaches $0.2506$ at $2.0k_0$. The task mean moves to a lower range. 

This trend indicates that PAR obtains its main predictive benefit within a limited rewiring budget, with additional edges mainly contributing further targeted shortage repair.

\textbf{Propagation depth.}
We further evaluate the sensitivity to the propagation depth $K$ in the pair-support term. To avoid the bias that larger $K$ accumulates more hop contributions under uniform weighting, this experiment uses the same MUTAG-GIN setting with a fixed decayed-support configuration. As shown in Fig.~\ref{fig:capacity}, $K=1$ gives the weakest structural repair, with $\Delta$Shortage of only $0.0218$. This indicates that a purely local propagation range is insufficient to capture pairwise communication shortage. Increasing $K$ to $2$ raises the task score to $0.8172$ and improves $\Delta$Shortage to $0.2389$, showing that a moderate propagation range provides a more effective support estimate. When $K$ increases to $3$ and $4$, $\Delta$Shortage continues to grow and reaches $0.3652$ and $0.6398$. The task means stay in a similar range under larger variance.

The result suggests that $K$ controls the scale at which PairAlign evaluates pairwise communication shortage. Task-relevant rewiring benefits are strongest when this scale matches the effective range of downstream message passing.

\section{Conclusion}
This work proposes PairAlign (PAR), an OT-guided graph rewiring framework that treats over-squashing as a pair-level communication shortage. PAR centers rewiring on node pairs whose structural demand is high but whose propagation support remains limited. The shortage score turns this pair-level bottleneck into an interpretable quantity as the ratio of current-graph support relative to original-graph demand, with theoretical consistency to the Jacobian-based shortage. PairAlign turns this diagnosis into an OT-guided allocation strategy for rewiring. Candidate additions are scored by current shortage reduction, with theoretical support for positive shortage relief on the rewired graph. OT-guided allocation then coordinates the limited budget across the shortage profile, prevents repair from concentrating on only a few targets, and obtains a coverage advantage over Greedy-Local under the coverage theory. Across standard graph benchmarks and different backbones, PAR shows consistent effectiveness, and its agreement with PER/TER supports pairwise shortage as a valid over-squashing bottleneck diagnostic.


How to explore task-aware demand modeling and explicit locality constraints for improving scalability on large graphs is our future work.

\bibliography{reference}

@inproceedings{abboud2022shortest,
  title={Shortest path networks for graph property prediction},
  author={Abboud, Ralph and Dimitrov, Radoslav and Ceylan, Ismail Ilkan},
  booktitle={Learning on graphs conference},
  year={2022},
  organization={PMLR}
}

@inproceedings{jamadandi2024spectral,
  title={Spectral Graph Pruning Against Over-Squashing and Over-Smoothing},
  author={Jamadandi, Adarsh and Rubio-Madrigal, Celia and Burkholz, Rebekka},
  booktitle={Advances in Neural Information Processing Systems},
  volume={37},
  year={2024}
}

@inproceedings{deac2022expander,
  title={Expander graph propagation},
  author={Deac, Andreea and Lackenby, Marc and Veli{\v{c}}kovi{\'c}, Petar},
  booktitle={Learning on Graphs Conference},
  pages={38--1},
  year={2022},
  organization={PMLR}
}

@article{bi2024make,
  title={Make heterophilic graphs better fit gnn: A graph rewiring approach},
  author={Bi, Wendong and Du, Lun and Fu, Qiang and Wang, Yanlin and Han, Shi and Zhang, Dongmei},
  journal={IEEE Transactions on Knowledge and Data Engineering},
  volume={36},
  number={12},
  pages={8744--8757},
  year={2024},
  publisher={IEEE}
}

@inproceedings{wilson2025cayley,
  title = 	 {Cayley Graph Propagation},
  author =       {Wilson, JJ and Bechler-Speicher, Maya and Veli{\v c}kovi{\' c}, Petar},
  booktitle = 	 {Proceedings of the Third Learning on Graphs Conference},
  year = 	 {2025},
  organization =    {PMLR},
}

@article{bose2025can,
  title={Can Graph Neural Networks Tackle Heterophily? Yes, With a Label-Guided Graph Rewiring Approach!},
  author={Bose, Kushal and Banerjee, Saptarshi and Das, Swagatam},
  journal={IEEE Transactions on Neural Networks and Learning Systems},
  year={2025},
  publisher={IEEE}
}

@inproceedings{bober2023rewiring,
  title={Rewiring networks for graph neural network training using discrete geometry},
  author={Bober, Jakub and Monod, Anthea and Saucan, Emil and Webster, Kevin N},
  booktitle={International Conference On Complex Networks And Their Applications},
  pages={225--236},
  year={2023},
  organization={Springer}
}

@inproceedings{banerjee2022oversquashing,
  title={Oversquashing in gnns through the lens of information contraction and graph expansion},
  author={Banerjee, Pradeep Kr and Karhadkar, Kedar and Wang, Yu Guang and Alon, Uri and Mont{\'u}far, Guido},
  booktitle={58th Annual Allerton Conference on Communication, Control, and Computing (Allerton)},
  pages={1--8},
  year={2022}
}

@inproceedings{arnaizdiffwire,
  title={DiffWire: Inductive Graph Rewiring via the Lov{\'a}sz Bound},
  author={Arnaiz-Rodr{\'\i}guez, Adri{\'a}n and Begga, Ahmed and Escolano, Francisco and Oliver, Nuria M},
  booktitle={Proceedings of the First Learning on Graphs Conference},
year={2022}
}

@book{villani2009optimal,
  title={Optimal transport: old and new},
  author={Villani, C{\'e}dric and others},
  volume={338},
  year={2009},
  publisher={Springer}
}

@inproceedings{cuturi2013sinkhorn,
  title={Sinkhorn Distances: Lightspeed Computation of Optimal Transport},
  author={Cuturi, Marco},
  booktitle={Advances in Neural Information Processing Systems},
  year={2013}
}

@article{akansha2025over,
  title={Over-squashing in graph neural networks: A comprehensive survey},
  author={Akansha, Singh},
  journal={Neurocomputing},
  volume={642},
  year={2025},
  publisher={Elsevier}
}

@inproceedings{chenrethinking,
  title={RETHINKING THE GOLD STANDARD: WHY DISCRETE CURVATURE FAILS TO FULLY CAPTURE OVER-SQUASHING IN GNNS?},
  author={Chen, Jialong and Deng, Bowen and Zheng, Zibin and Chen, Chuan},
  booktitle={International Conference on Learning Representations},
  year={2026}
}

@article{shen2024graph,
  title={Graph rewiring and preprocessing for graph neural networks based on effective resistance},
  author={Shen, Xu and Lio, Pietro and Yang, Lintao and Yuan, Ru and Zhang, Yuyang and Peng, Chengbin},
  journal={IEEE Transactions on Knowledge and Data Engineering},
  volume={36},
  number={11},
  pages={6330--6343},
  year={2024},
  publisher={IEEE}
}

@inproceedings{altschuler2017nearlinear,
  title={Near-Linear Time Approximation Algorithms for Optimal Transport via Sinkhorn Iteration},
  author={Altschuler, Jason and Weed, Jonathan and Rigollet, Philippe},
  booktitle={Advances in Neural Information Processing Systems},
  year={2017}
}

@inproceedings{torieffectiveness,
  title={The Effectiveness of Curvature-Based Rewiring and the Role of Hyperparameters in GNNs Revisited},
  author={Tori, Floriano and Holst, Vincent and Ginis, Vincent},
  booktitle={International Conference on Learning Representations},
  year={2025}
}

@inproceedings{platonov2023critical,
  title={A Critical Look at the Evaluation of GNNs under Heterophily: Are We Really Making Progress?},
  author={Platonov, Oleg and Kuznedelev, Denis and Diskin, Mikhail and Babenko, Artem and Prokhorenkova, Liudmila},
  booktitle={Advances in Neural Information Processing Systems},
  year={2023}
}

@inproceedings{velickovic2018graph,
  title={Graph Attention Networks},
  author={Veli{\v{c}}kovi{\'c}, Petar and Cucurull, Guillem and Casanova, Arantxa and Romero, Adriana and Li{\`o}, Pietro and Bengio, Yoshua},
  booktitle={International Conference on Learning Representations},
  year={2018}
}

@inproceedings{black2023understanding,
  title={Understanding oversquashing in gnns through the lens of effective resistance},
  author={Black, Mitchell and Wan, Zhengchao and Nayyeri, Amir and Wang, Yusu},
  booktitle={International Conference on Machine Learning},
  pages={2528--2547},
  year={2023},
  organization={PMLR}
}

@inproceedings{gilmer2017neural,
  title={Neural message passing for quantum chemistry},
  author={Gilmer, Justin and Schoenholz, Samuel S and Riley, Patrick F and Vinyals, Oriol and Dahl, George E},
  booktitle={International conference on machine learning},
  pages={1263--1272},
  year={2017},
  organization={Pmlr}
}

@article{kipf2016semi,
  title={Semi-supervised classification with graph convolutional networks},
  author={Kipf, Thomas N and Welling, Max},
  journal={arXiv preprint arXiv:1609.02907},
  year={2016}
}

@inproceedings{alonbottleneck,
  title={On the Bottleneck of Graph Neural Networks and its Practical Implications},
  author={Alon, Uri},
  booktitle={International Conference on Learning Representations},
year={2021}
}

@article{gabrielsson2023rewiring,
  title={Rewiring with positional encodings for graph neural networks},
  author={Gabrielsson, Rickard Br{\"u}el and Yurochkin, Mikhail and Solomon, Justin},
  journal={Transactions on Machine Learning Research},
  year={2023}
}

@inproceedings{nguyen2023revisiting,
  title={Revisiting over-smoothing and over-squashing using ollivier-ricci curvature},
  author={Nguyen, Khang and Hieu, Nong Minh and Nguyen, Vinh Duc and Ho, Nhat and Osher, Stanley and Nguyen, Tan Minh},
  booktitle={International Conference on Machine Learning},
  pages={25956--25979},
  year={2023},
  organization={PMLR}
}

@inproceedings{toppingunderstanding,
  title={Understanding over-squashing and bottlenecks on graphs via curvature},
  author={Topping, Jake and Di Giovanni, Francesco and Chamberlain, Benjamin Paul and Dong, Xiaowen and Bronstein, Michael M},
  booktitle={International Conference on Learning Representations},
year={2022}
}

@inproceedings{karhadkar2023fosr,
  title={FoSR: First-order spectral rewiring for addressing oversquashing in GNNs},
  author={Karhadkar, Kedar and Banerjee, Pradeep and Montufar, Guido},
  booktitle={International Conference on Learning Representations},
  year={2023}
}

@inproceedings{barberolocality,
  title={Locality-Aware Graph Rewiring in GNNs},
  author={Barbero, Federico and Velingker, Ameya and Saberi, Amin and Bronstein, Michael M and Di Giovanni, Francesco},
  booktitle={International Conference on Learning Representations},
 year={2024}
}

@inproceedings{liangmitigating,
  title={Mitigating Over-Squashing in Graph Neural Networks by Spectrum-Preserving Sparsification},
  author={Liang, Langzhang and Bu, Fanchen and Song, Zixing and Xu, Zenglin and Pan, Shirui and Shin, Kijung},
  booktitle={International Conference on Machine Learning},
year={2025}
}

@inproceedings{rubiognns,
  title={GNNs Getting ComFy: Community and Feature Similarity Guided Rewiring},
  author={Rubio-Madrigal, Celia and Jamadandi, Adarsh and Burkholz, Rebekka},
  booktitle={International Conference on Learning Representations},
year={2025}
}

@inproceedings{linkerhagnerjoint,
  title={Joint Graph Rewiring and Feature Denoising via Spectral Resonance},
  author={Linkerh{\"a}gner, Jonas and Shi, Cheng and Dokmani{\'c}, Ivan},
  booktitle={International Conference on Learning Representations},
  year={2025}
}

@inproceedings{yang2016revisiting,
  title={Revisiting semi-supervised learning with graph embeddings},
  author={Yang, Zhilin and Cohen, William and Salakhudinov, Ruslan},
  booktitle={International conference on machine learning},
  pages={40--48},
  year={2016},
  organization={PMLR}
}

@inproceedings{pei2020geom,
  title={GEOM-GCN: GEOMETRIC GRAPH CONVOLUTIONAL NETWORKS},
  author={Pei, Hongbin and Wei, Bingzhe and Chang, Kevin Chen Chuan and Lei, Yu and Yang, Bo},
  booktitle={International Conference on Learning Representations},
  year={2020}
}

@article{rozemberczki2021multi,
  title={Multi-scale attributed node embedding},
  author={Rozemberczki, Benedek and Allen, Carl and Sarkar, Rik},
  journal={Journal of Complex Networks},
  volume={9},
  number={2},
  pages={1-22},
  year={2021}
}

@article{bengio2013estimating,
  title={Estimating or propagating gradients through stochastic neurons for conditional computation},
  author={Bengio, Yoshua and L{\'e}onard, Nicholas and Courville, Aaron},
  journal={arXiv preprint arXiv:1308.3432},
  year={2013}
}

@article{morris2020tudataset,
  title={Tudataset: A collection of benchmark datasets for learning with graphs},
  author={Morris, Christopher and Kriege, Nils M and Bause, Franka and Kersting, Kristian and Mutzel, Petra and Neumann, Marion},
  journal={arXiv preprint arXiv:2007.08663},
  year={2020}
}

@article{xu2018powerful,
  title={How powerful are graph neural networks?},
  author={Xu, Keyulu and Hu, Weihua and Leskovec, Jure and Jegelka, Stefanie},
  journal={arXiv preprint arXiv:1810.00826},
  year={2018}
}
\end{document}